\documentclass{article}

\usepackage{subcaption}
 \usepackage[preprint]{neurips_2025}

\usepackage[utf8]{inputenc} 
\usepackage[T1]{fontenc}    
\usepackage{hyperref}       
\usepackage{url}            
\usepackage{booktabs}       
\usepackage{amsfonts}       
\usepackage{nicefrac}       
\usepackage{microtype}      
\usepackage{xcolor}

\usepackage[utf8]{inputenc} 
\usepackage[T1]{fontenc}    
\usepackage{hyperref}       
\usepackage{url}            
\usepackage{booktabs}       
\usepackage{amsfonts}       
\usepackage{nicefrac}       
\usepackage{microtype}      
\usepackage{multirow}
\usepackage{graphicx}
\usepackage{amsmath}
\usepackage{adjustbox}
\usepackage{pifont}
\newcommand{\cmark}{\ding{51}}  
\newcommand{\xmark}{\ding{55}}  
\usepackage{colortbl}
\usepackage{enumitem}
\usepackage{makecell}
\usepackage{fontawesome}

\definecolor{myteal}{RGB}{128, 0, 128}
\definecolor{top1}{RGB}{255,200,200} 
\definecolor{top2}{RGB}{255,230,190}      
\definecolor{top3}{RGB}{255,255,200}      

\usepackage[most,breakable]{tcolorbox}
\usepackage{hyperref}

\tcbset{
  rqlist/.style={
    enhanced,
    colback=purple!5,
    colframe=purple!50!black,
    boxrule=0.8pt,
    arc=3pt,
    left=5pt,
    right=5pt,
    top=5pt,
    bottom=5pt,
    fonttitle=\bfseries,
  }
}

\DeclareRobustCommand{\observationbox}[2][blue!5]{%
\begin{tcolorbox}[   
        breakable,
        left=0pt,
        right=0pt,
        top=0pt,
        bottom=0pt,
        colback=#1,
        colframe=#1,
        width=\columnwidth,
        arc=1pt,outer arc=1pt,
        ]
        #2
\end{tcolorbox}
}

\title{E3D-Bench: A Benchmark for \\ End-to-End 3D Geometric Foundation Models}

\author{%
\textbf{Wenyan Cong}$^{1}$, 
\textbf{Yiqing Liang}$^{2}$, 
\textbf{Yancheng Zhang}$^{3}$, 
\textbf{Ziyi Yang}$^{1}$, \\
\textbf{Yan Wang}$^{4}$, 
\textbf{Boris Ivanovic}$^{4}$, 
\textbf{Marco Pavone}$^{4,5}$, 
\textbf{Chen Chen}$^{3}$, 
\textbf{Zhangyang Wang}$^{1}$\thanks{co-corresponding authors},  
\textbf{Zhiwen Fan}$^{1*}$\\
  $^1$University of Texas at Austin\quad
  $^2$Brown University\quad 
  $^3$University of Central Florida\quad \\
  $^4$NVIDIA Research\quad
  $^5$Stanford University \\
\textbf{Website:}\href{https://e3dbench.github.io/}{\faGlobe\ \textcolor{myteal}{\texttt{https://e3dbench.github.io/}}} \\
\textbf{Code:} \href{https://github.com/VITA-Group/E3D-Bench}{\faGithub\ \textcolor{myteal}{\texttt{https://github.com/VITA-Group/E3D-Bench}}}
}

\begin{document}
\maketitle

\vspace{-20pt}
\begin{figure}[ht]
\centering
\includegraphics[width=\linewidth]{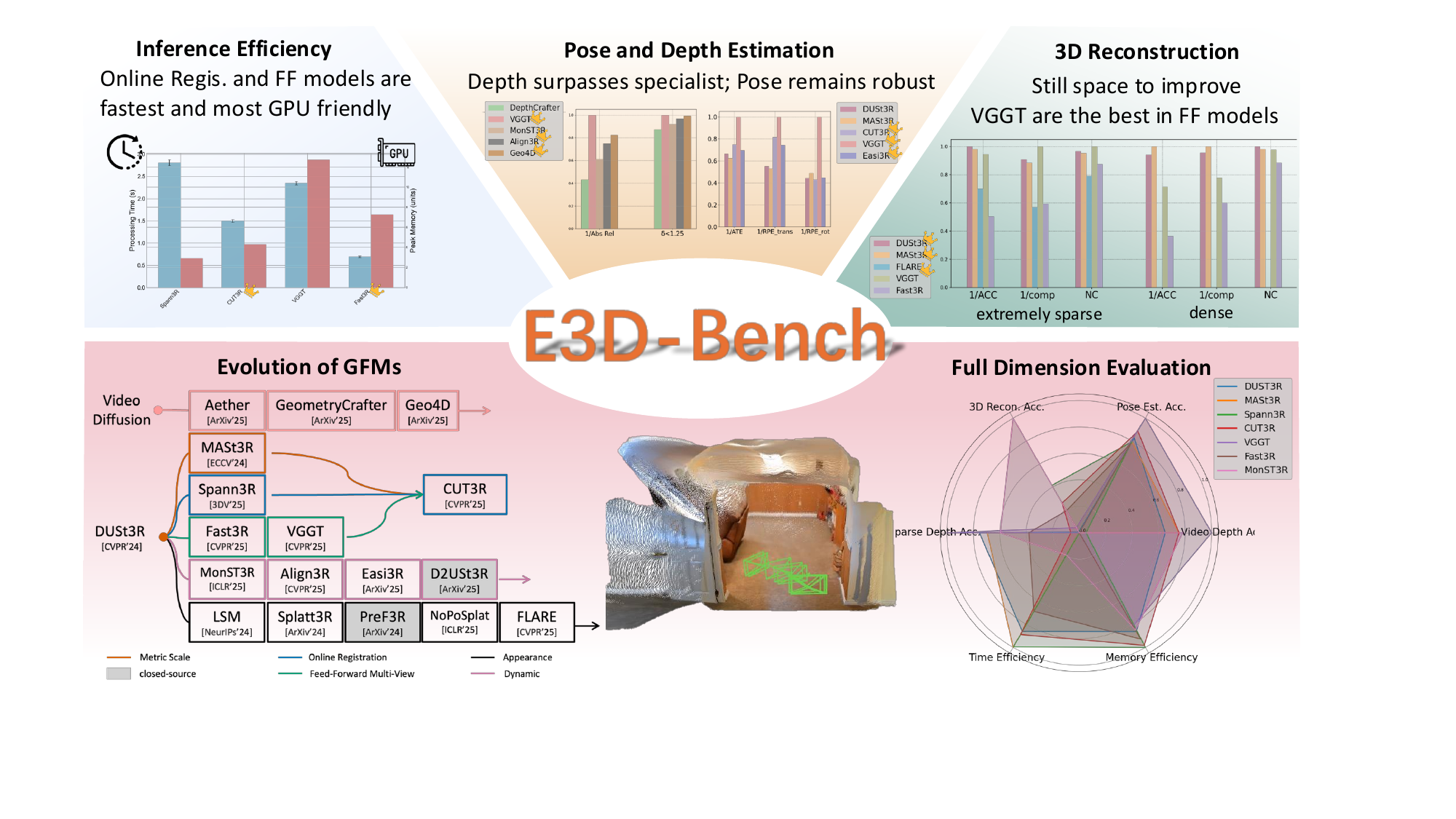}
\vspace{-20pt}
\caption{\small \textbf{E3D-Bench} evaluates 16 recent 3D Geometric Foundation Models (GFMs), spanning two major architectural families: feed-forward ViT-based and diffusion-based models, on both effectiveness and efficiency. For clarity, we visualize only the top methods per task. All metrics in bin chart except inference time, are averaged per scene, normalized, and converted to a "higher-is-better" scale for consistent comparison.}
\label{fig:teaser}
\end{figure}

\begin{abstract}

Spatial intelligence, encompassing 3D reconstruction, perception, and reasoning, is fundamental to applications such as robotics, aerial imaging, and extended reality. 
A key enabler is the real‑time, accurate estimation of core 3D attributes (camera parameters, point clouds, depth maps, and 3D point tracks) from unstructured or streaming imagery. 
Inspired by the success of large foundation models in language and 2D vision, a new class of end‑to‑end 3D geometric foundation models (GFMs) has emerged, directly predicting dense 3D representations in a single feed‑forward pass, 
eliminating the need for slow or unavailable precomputed camera parameters.

Since late 2023, the field has exploded with diverse variants, but systematic evaluation is lacking.
In this work, we present the first comprehensive benchmark for 3D GFMs, covering five core tasks: sparse‑view depth estimation, video depth estimation, 3D reconstruction, multi‑view pose estimation, novel view synthesis, and spanning both standard and challenging out‑of‑distribution datasets. 
Our standardized toolkit automates dataset handling, evaluation protocols, and metric computation to ensure fair, reproducible comparisons. We evaluate 16 state‑of‑the‑art GFMs, revealing their strengths and limitations across tasks and domains, and derive key insights to guide future model scaling and optimization. All code, evaluation scripts, and processed data will be publicly released to accelerate research in 3D spatial intelligence.

\end{abstract}
\section{Introduction}\label{sec:intro}
3D reconstruction, perception, and reasoning are foundational steps in spatial intelligence, enabling downstream applications like robotic perception, extended reality, and embodied agents to interact reliably with the physical world. A critical prerequisite is \textbf{\textit{fast}}, \textbf{\textit{accurate}} estimation of core 3D attributes (camera parameters, point clouds, depth maps, and tracked 3D points) from unstructured or streaming images. Traditional multi‑stage pipelines—e.g., Structure‑from‑Motion~\cite{schoenberger2016mvs,yao2018mvsnet,gu2020cascade} or optimization‑based dense SLAM~\cite{kerl2013dense,whelan2016elasticfusion,Tateno_2017_CVPR}—incur high computational cost and maintenance complexity, further limiting their real‑time deployment on latency-constrained scenarios such as robotic perception. They also struggle under sparse viewpoints, dynamic content, or non‑ideal capture settings, leading to failures in open‑world, latency‑sensitive scenarios~\cite{Luo-VideoDepth,lasinger2019towards}. 



\noindent\textbf{Rise of 3D Geometric Foundation Models (GFMs).} Driven by the success of large foundation models in language~\cite{achiam2023gpt} and 2D vision~\cite{kirillov2023segment, pagnoni2024byte},
a new wave of generalizable 3D GFMs has emerged with end-to-end training and inference capabilities.
These GFMs predict dense 3D representations (e.g., point maps and depth maps) directly from multi-view images,
eliminating the need for slow-to-obtain or unavailable pre‑computed camera parameters in open-world scenarios~\cite{cong2025videolifter}.
Moreover, they unify feature extraction, matching, and geometry verification in an end-to-end single feed‑forward pass with inference speeds in seconds, instead of traditional multi‑stage reconstruction pipelines.

\noindent\textbf{Rapid Evolution of GFMs.} Since late 2023, this paradigm has seen explosive growth (See Tab.~\ref{tab:method-overview}), reflecting a broader shift toward scalable, unified architectures for dense 3D understanding, empowering end‑to‑end \textit{reconstruction} and \textit{perception} in next‑generation spatially aware systems.
As illustrated in Fig.~\ref{fig:teaser}, recent GFMs adopt transformer-based architectures, and primarily follow two major architectural branches: 
\textbf{\ding{192}~Feed-forward models} (e.g., DUSt3R~\cite{dust3r} and its variants~\cite{mast3r, Yang_2025_Fast3R, zhang2025monstr, wang2025vggt, fan2024largespatialmodelendtoend, wang2025cut3r}), which predict geometry in a single pass, and \textbf{\ding{193}~Diffusion-based models} (e.g., Aether~\cite{team2025aether}, Geo4D~\cite{Geo4D}), which iteratively refine geometry through a denoising process.
Notably, models derived from DUSt3R have diversified into memory-augmented designs, multi-view fusion strategies, and appearance-aware variants---each striking a unique balance among reconstruction quality, scalability, and inference efficiency to suit diverse deployment scenarios.



\noindent\textbf{A Timely Assessment of GFMs.} With the rapid proliferation of 3D GFMs, we ask:
\\[4pt]
\noindent\colorbox{yellow!20}{Q \ding{202}{~Can GFMs serve as an \textit{\textbf{effective}} and \textit{\textbf{robust}} foundation for diverse 3D tasks and scenarios?}}
  \\[4pt]
\noindent\colorbox{blue!10}{Q \ding{203}{~Can GFMs serve as an \textit{\textbf{efficient}} foundation, especially for latency‑constrained 3D applications?}}
\\[4pt]
These questions are critical because the final desired 3D GFMs for real‑world 3D tasks must: \ding{172} Deliver high‑quality 3D predictions, whether via end‑to‑end geometry output or through decomposed subtasks (e.g., depth and pose estimation).
\ding{173} Generalize across diverse capture settings such as indoor vs.\ outdoor environments, sparse vs.\ dense view settings, ground-based vs.\ aerial perspectives, and static vs.\ dynamic scenes.
\ding{174} Able to operate under strict latency constraints, especially for real-time applications.


Hence, in the era of foundation models, we present the first systematic benchmark for 3D geometric foundation models, \textit{aiming to uncover key trends that will guide future scaling and optimization.}

\noindent\textbf{First Comprehensive Benchmark for End-to-end 3D GFMs.}  
As summarized in Tab.~\ref{tab:method-overview}, we evaluate a broad spectrum of GFMs, from pioneering feed‑forward backbones (DUSt3R~\cite{dust3r}, MASt3R~\cite{mast3r}) and their monocular‑video adaptations (MonST3R~\cite{zhang2025monstr}) to multi‑view variants (VGGT~\cite{wang2025vggt}) and diffusion‑based models (Geo4D~\cite{Geo4D}), among others. 
\textit{To rigorously assess \textbf{effectiveness}}, we cover five core tasks: (a) Sparse‑view depth estimation (b) Video‑based depth estimation (c) Multi‑view 3D reconstruction (both sparse and dense regimes) (d) Multi‑view relative pose estimation (e) Novel view synthesis.  
\textit{To probe \textbf{robustness}}, 
we incorporate datasets spanning \textit{diverse domains} beyond standard indoor scenes, including drone footage, dynamic real‑world sequences, and air‑ground paired views.
Finally, we \textit{benchmark latency and memory usage} to evaluate each model’s readiness for latency‑constrained, real‑time deployment.  

\begin{table}[ht]
\centering
\scriptsize
\vspace{-5pt}
\caption{\small \textbf{Overview of evaluated end-to-end 3D geometric foundation models.}
Methods are grouped by input type. \cmark~denotes support for metric scale; \xmark~denotes normalized scale. FT indicates fine-tuning on a pretrained DUSt3R. Abbreviations are defined below. Confidence outputs are omitted for clarity.}
\label{tab:method-overview}
\resizebox{\linewidth}{!}{
\begin{tabular}{lccccccc}
\toprule
\makecell{\textbf{Method}} & 
\makecell{\textbf{ArXiv} \\ \textbf{Date}} & 
\makecell{\textbf{Publication}} & 
\makecell{\textbf{Output} \\ \textbf{Modality}} & 
\makecell{\textbf{Metric} \\ \textbf{Scale}} & 
\makecell{\textbf{Multi-View} \\ \textbf{Handling}} & 
\makecell{\textbf{Backbone}} & 
\makecell{\textbf{\# Train} \\ \textbf{Datasets}} \\
\midrule
\multicolumn{8}{l}{\cellcolor{blue!5} \raggedright \textbf{Pair of Images}} \\
DUSt3R~\cite{dust3r}    & 2312 & CVPR'24 & PM            & \xmark & GA & ViT & 9\\
MASt3R~\cite{mast3r}    & 2406& ECCV'24 & PM, M         & \cmark & GA & ViT & 14\\
LSM~\cite{fan2024largespatialmodelendtoend}      & 2410& NeurIPS'24 & PM, 3DGS, S   & \xmark & GA & ViT & 2\\
MonST3R~\cite{zhang2025monstr}   & 2410& ICLR'25 & PM            & \xmark & GA & ViT & FT 4\\
NoPoSplat~\cite{ye2024no}    & 2410& ICLR'25 & 3DGS            & \xmark & - & ViT & 1-2\\
Align3R~\cite{lu2024align3r}   & 2412& CVPR'25 & PM            & \xmark & GA & ViT & FT 5\\
Splatt3R~\cite{smart2024splatt3r}    & 2408& ArXiv'24 & PM, M, 3DGS            & \xmark & GA & ViT & 1\\
Easi3R~\cite{chen2025easi3r}    & 2503& ArXiv'25 & PM            & \xmark & GA & ViT & 0\\

\midrule
\multicolumn{8}{l}{\cellcolor{blue!5} \raggedright \textbf{Image Sequences}} \\
Spann3R~\cite{spann3r}   & 2408& 3DV'25 & PM      & \xmark & OR & ViT & 6\\
CUT3R~\cite{wang2025cut3r}     & 2501& CVPR'25 & PM, PM$_c$, Pose      & \cmark & OR & ViT & 32\\
Aether~\cite{team2025aether}    & 2503& ArXiv'25 & PM      & \xmark & FF & Diffusion & 1\\
Geo4D~\cite{Geo4D}      & 2504& ArXiv'25 & PM, Ray      & \xmark & GA & Diffusion & 5\\
GeometryCrafter~\cite{xu2025geometrycrafter}     & 2504& ArXiv'25 & PM      & \xmark & GA & Diffusion & 14\\
\midrule
\multicolumn{8}{l}{\cellcolor{blue!5} \raggedright \textbf{Multi-view Images}} \\
Fast3R~\cite{Yang_2025_Fast3R}    & 2501& CVPR'25 & PM, PM$_c$, Pose      & \xmark & FF & ViT & 6\\
VGGT~\cite{wang2025vggt}      & 2503& CVPR'25 & PM, PM$_c$, Pose, M   & \xmark & FF & ViT & 9\\
\midrule
\multicolumn{8}{l}{\cellcolor{blue!5} \raggedright \textbf{Sparse-view Images}} \\
FLARE~\cite{zhang2025FLARE}     & 2502& CVPR'25 & PM, 3DGS, Pose & \xmark & FF & ViT & 8\\
\bottomrule
\end{tabular}}

\vspace{0.5em}
\raggedright
{\tiny
\textbf{Output Modality:} PM = Pointmap (world coordinate), PM$_c$ = Pointmap (camera coordinate), M = Matching, Pose = Camera Pose, Ray = Ray Map, 3DGS = 3D Gaussian Splatting, S = Semantics \\
\textbf{Multi-View Handling:} GA = Global Alignment, OR = Online Registration, FF = Feed-Forward, – = Not Supported \\
\textbf{Backbone:} All GFMs adopt transformer-based architectures. For clarity, ViT = Feed-Forward Transformer; Diffusion = Transformer-based Iterative Denoising
}
\vspace{-10pt}
\end{table}



\textbf{Key Findings.} We summarize findings on 3D GFMs' effectiveness, robustness, and efficiency below:

\begin{description}[leftmargin=20pt,labelindent=10pt]
    \item[1. Impact of Task Difficulty (\S\ref{sec:rq1})]: 
    Current GFMs excel on simpler sub‑tasks but struggle as complexity grows:   \ding{172} Pair‑view geometry inference outperforms true multi‑view scenarios.  \ding{173} Single‑attribute predictions (depth or pose) are more reliable than full 3D scene reconstruction.  \ding{174} Relative metrics (e.g., relative depth) are more accurate than absolute, metric‑scale outputs.  
    \item[2. Impact of Data Domains (\S\ref{sec:rq2})]: GFMs generalize well on in-domain and out-of-domain data regimes, but degrade sharply under extreme distribution shifts, highlighting the need for more diverse training data.
    \item[3. Insights on Model Architecture (\S\ref{sec:rq3})]: No single backbone type (feed-forward ViT or diffusion) dominates; architecture choice should align with task needs. Moreover, leveraging strong 2D feature extractors (e.g., DINO~\cite{dino}) substantially boosts 3D performance.
    \item[4. Efficiency Analysis (\S\ref{sec:rq4})]: Inference latency has improved, but modern GFMs still require tens of seconds to process hundreds of views, highlighting that efficiency—alongside accuracy—is essential for real‑time deployment and for the future evolution of GFMs.
\end{description}

Looking ahead, despite promising progress, current GFMs still fall short of delivering a plug-and-play solution for end-to-end 3D attribute prediction in real-world spatial intelligence applications. Challenges such as metric-scale accuracy, generalization to diverse capture settings, and efficiency under latency constraints remain open.
To facilitate progress in this direction, we present what is, to the best of our knowledge, the first systematic benchmark that spans a comprehensive set of 3D tasks and diverse real-world scenarios. We release all code and evaluation tools to support reproducibility and catalyze further research on robust, generalizable, and efficient 3D geometric foundation models.

\section{Background and Related Work}

Due to space constraints, we provide only a brief overview of relevant work in Tab.~\ref{tab:method-overview}. Descriptions of evaluated models and datasets appear in the next sections, and a comprehensive discussion of related literature is included in Appendix~\ref{apx:related_work}.

\section{Benchmark Design and Results}

To address the questions posed in Section~\ref{sec:intro} regarding the \emph{effectiveness}, \emph{robustness}, and \emph{efficiency} of end‑to‑end 3D GFMs, we select five representative tasks and assemble a diverse suite of datasets, ranging from standard indoor scenes to challenging scenarios such as drone‑captured footage.  

\subsection{Sparse-View Depth Estimation}\label{sec:sparse_depth}

\textbf{Task:} \textit{This task aims to predict the per-pixel depth maps, given a sparse view setting (views with minimal/no overlap).}
This is a relatively underexplored task, with only a few recent end-to-end 3D GFMs, such as DUSt3R~\cite{dust3r} and Fast3R~\cite{Yang_2025_Fast3R}, explicitly adopting it in evaluation. 
However, \textit{this task is critical for real-world deployment}, particularly in situations where sparse and unstructured image collections are the norm (e.g., internet photo collections of landmarks). 
In such cases, traditional pipelines like COLMAP~\cite{schoenberger2016sfm, schoenberger2016mvs} often fail due to insufficient keypoint matches or poor overlap. 
Although most end-to-end GFMs are not explicitly trained for depth estimation,
end‑to‑end GFMs naturally produce dense, stereo‑consistent depth by projecting their predicted point maps and extracting the z‑coordinate relative to each view. 
Assessing performance on sparse‑view depth estimation, therefore, offers key insight into a model’s ability to generalize and reason about 3D structure under severely limited visual input.


\textbf{Dataset:} Five widely used datasets: DTU~\cite{dtu}, ETH3D~\cite{eth3d}, KITTI~\cite{kitti}, Tanks and Temples~\cite{tanks}, and ScanNet~\cite{scannet} that collectively cover diverse domains including object-centric scenes, large-scale indoor/outdoor scenes, and street driving scenes, and span diverse scene scales from 0.2m to 85m.

\textbf{Evaluation Metrics:}
Two widely used metrics to evaluate depth estimation performance: 
Absolute Relative Error (AbsRel) measures the average relative error between model output and ground truth;  the Inlier Ratio $\delta<1.03$~\cite{uhrig2017sparsity, eigen2014depth} captures the percentage of pixels within 3\% relative error.

\textbf{Evaluation Protocol:}  
We extract depth maps from the z‑coordinate of predicted point maps, or use confidence-weighted averaging as in DUSt3R for views with multiple predicted depths.
We test both normalized and metric-scale models, where we evaluate the former using median depth scaling, and the latter under their raw outputs and an extra median-aligned setting for fair comparison.
All predictions are upsampled to full resolution before evaluation.
We select quasi-optimal source views to reduce source view selection bias, following~\cite{schroppel2022benchmark}.
More details can be found in Appendix~\ref{apx:sparse_depth}.

\textbf{Quantitative Results:} As shown in
Tab.~\ref{tab:sparse_depth}: 1) Most GFMs perform on par with or better than specialized baselines like~\cite{schroppel2022benchmark}, demonstrating strong robustness in this extreme case.
2) However, in the metric-scale setting, MASt3R~\cite{mast3r} and CUT3R~\cite{wang2025cut3r} fail to meet the strict threshold (e.g., $<3\%$ error), highlighting the difficulty of accurate scale recovery under sparse inputs.

\begin{table}[ht]
\centering
\scriptsize
\vspace{-5pt}
\caption{\small \textbf{Comparison on Sparse-View Depth Estimation}. We report Absolute Relative Error (Abs Rel ↓) and $\delta{<}1.03$ accuracy (↑). Since LSM~\cite{fan2024largespatialmodelendtoend} builds on top of DUSt3R without modifying its weights, their performance is nearly identical. For clarity and conciseness, we report them jointly as DUSt3R/LSM by default.}
\label{tab:sparse_depth}
\resizebox{0.98\linewidth}{!}{
\begin{tabular}{llcccccccccc}
\toprule
\multirow{2}{*}{\textbf{Scale}} & \multirow{2}{*}{\textbf{Method}} 
& \multicolumn{2}{c}{\cellcolor{gray!20}\textbf{DTU}} 
& \multicolumn{2}{c}{\cellcolor{blue!15}\textbf{ScanNet}} 
& \multicolumn{2}{c}{\cellcolor{green!15}\textbf{KITTI}} 
& \multicolumn{2}{c}{\cellcolor{pink!20}\textbf{ETH3D}} 
& \multicolumn{2}{c}{\cellcolor{pink!20}\textbf{T\&T}} \\
&& \cellcolor{gray!20}AbsRel ↓ & \cellcolor{gray!20}$\delta{<}1.03$ 
↑& \cellcolor{blue!15}AbsRel ↓& \cellcolor{blue!15}$\delta{<}1.03$ ↑  
& \cellcolor{green!15}AbsRel ↓& \cellcolor{green!15}$\delta{<}1.03$ ↑ 
& \cellcolor{pink!20}AbsRel ↓& \cellcolor{pink!20}$\delta{<}1.03$ ↑ 
& \cellcolor{pink!20}AbsRel ↓& \cellcolor{pink!20}$\delta{<}1.03$ ↑ \\
\midrule
\multirow{8}{*}{\rotatebox{90}{Normalized}}
&Robust MVD & \cellcolor{top2} 2.490 & \cellcolor{top2} 80.056 & 7.468 & 35.651 & \cellcolor{top2} 9.419 & 30.505 & 9.302 & 42.909 & 6.379 & 58.409 \\\cmidrule(lr){2-12}
&DUSt3R/LSM & \cellcolor{top3} 2.741 & \cellcolor{top3} 75.685 & \cellcolor{top2} 4.732 & \cellcolor{top2} 61.337 & \cellcolor{top1} 9.113 & 39.495 & \cellcolor{top3} 3.132 & \cellcolor{top3} 74.851 & \cellcolor{top3} 3.106 & \cellcolor{top3} 77.033\\
&MASt3R & 3.343 & 68.301 & 5.949 & \cellcolor{top3} 54.516 & 9.542 & \cellcolor{top1} 46.805 & \cellcolor{top2} 2.471 & \cellcolor{top2} 81.291 & \cellcolor{top2} 2.381 & \cellcolor{top2} 82.262\\
&Spann3R & 6.431 & 38.339 & 7.779 & 33.713 & 10.195 & 30.858 & 5.121 & 54.708 & 5.580 & 52.812\\
&CUT3R & 6.200 & 47.421 & 8.231 & 39.464 & 23.849 & 12.087 & 5.224 & 59.864 & 4.594 & 56.773\\
&VGGT  & \cellcolor{top1} 1.085 & \cellcolor{top1} 94.305 & \cellcolor{top1} 4.386 & \cellcolor{top1} 64.968 & \cellcolor{top3} 9.436 &  \cellcolor{top2}41.309 & \cellcolor{top1} 1.782 & \cellcolor{top1} 86.337 & \cellcolor{top1} 2.075 & \cellcolor{top1} 85.174\\
&Fast3R & 3.940 & 62.120 & 6.271 & 50.283 & 13.390 & 26.734 & 4.692 & 62.663 & 4.423 & 64.873 \\
&MonST3R  & 5.346 & 67.977 & \cellcolor{top3} 5.557 & 53.309 & 10.191 & \cellcolor{top3} 40.274 & 3.368 & 72.624 & 3.289 & 72.491 \\
\cmidrule(lr){1-12}
\multirow{3}{*}{Metric} 
& Robust MVD & 2.242 & 84.574 & 8.016 & 35.924 & 10.846 & 25.534 & 10.944 & 35.526 & 6.982 & 60.643 \\ \cmidrule(lr){2-12}
&MASt3R & 84.904 & 0.000 & 93.584 & 0.000 & 99.069 & 0.000 & 97.021 & 0.000 & 98.234 & 0.000 \\
& CUT3R & 84.904 & 0.000 & 93.584 & 0.000 & 99.069 & 0.000 & 97.022 & 0.000 & 98.234 & 0.000 \\
\bottomrule
\end{tabular}
}
{\tiny
\begin{tabular}{@{}l@{}}
\colorbox{gray!20}{\strut\textbf{Gray}}: Object-Centric \quad
\colorbox{blue!15}{\strut\textbf{Blue}}: Indoor Scene \quad
\colorbox{green!15}{\strut\textbf{Green}}: Outdoor Scene \quad
\colorbox{pink!20}{\strut\textbf{Pink}}: Mixed Scene \quad
\end{tabular}
}
\vspace{-10pt}
\end{table}

\subsection{Video Depth Estimation}
\label{sec:video_depth}
\textbf{Task:}
\textit{This task evaluates producing temporally consistent depth maps from monocular video.}
Coherent predictions across consecutive frames, despite motion blur, occlusions, or dynamic objects, are critical for applications such as robot navigation, SLAM, and AR. Methods like MonST3R~\cite{zhang2025monstr} and Geo4D~\cite{Geo4D} are explicitly designed for dynamic scenes, but most existing GFMs have not been tested in temporally structured settings. We benchmark both dynamic-aware and static-scene models on video sequences, applying the latter to assess their zero-shot generalization to real-world dynamics. This enables analysis of not just single-frame accuracy but also the stability and robustness of depth predictions over time, key requirements for video-driven 3D perception systems.

\textbf{Dataset:} Six video datasets: 1) Bonn~\cite{bonn} and 2) TUM Dynamics~\cite{tumdynamic}: indoor scenes with human motion and interaction, 3) KITTI~\cite{kitti}: dynamic street-driving scenes, 4) Sintel~\cite{sintel}: both indoor and outdoor scenes with complex motion, occlusion, and lighting changes, 5) PointOdyssey~\cite{pointodyssey} validation split: large dynamic motion with frequent occlusions, 6) Syndrone~\cite{syndrone}: outdoor drone footage with rapid viewpoint changes and wide-baseline shifts. Together, these datasets cover indoor, outdoor, and aerial, synthetic and real, and static and highly dynamic settings.

\textbf{Evaluation Protocol and Metrics:} As in Sec.~\ref{sec:sparse_depth}, we report \textit{Absolute Relative Error} (AbsRel) and \textit{Inlier Ratio} \( \delta \)~\cite{uhrig2017sparsity, eigen2014depth} with a higher threshold 1.25, and evaluate both normalized and metric-scale models.
Models do not have access to camera intrinsics or ground-truth poses during inference.
More details can be found in Appendix~\ref{apx:video_depth}.

\textbf{Quantitative Results:} As shown in Tab.~\ref{tab:video_depth}: 1) Surprisingly, most GFMs outperform methods specifically designed for recent SOTA video depth estimation~\cite{video_depth_anything,yang2024depthanyvideo, ke2025marigold,hu2024depthcrafter}.
2) VGGT~\cite{wang2025vggt} consistently achieves the best performance across all domains.
3) Geo4D~\cite{Geo4D} is the next strongest overall, while other diffusion-based models perform less competitively. Among DUSt3R-style models, MonST3R~\cite{zhang2025monstr} and Align3R~\cite{lu2024align3r}, both fine-tuned on video data, achieve the second best results.
4) Estimating depth in metric-scale remains more challenging; we observe significant improvements from MASt3R~\cite{mast3r} to CUT3R~\cite{wang2025cut3r}, reflecting better scale recovery and calibration. It should be noted that for Geo4D~\cite{Geo4D}, with the input sequences containing fewer than 16 frames, we pad them to 16 frames by repeating the last frame of the input video. For Aether~\cite{team2025aether}, we use its default one-shot output of 41 frames for evaluation. The same settings are applied in all subsequent experiments.

\begin{table}[ht]
\centering
\scriptsize
\vspace{-5pt}
\caption{\textbf{Comparison on Video Depth Estimation.} We report Abs Rel (↓) and $\delta{<}1.25$ (↑) cross diverse dataset domains.}
\label{tab:video_depth}
\resizebox{\linewidth}{!}{
\begin{tabular}{l l cc cc cc cc cc cc cc}
\toprule
\multirow{2}{*}{\textbf{Scale}} & \multirow{2}{*}{\textbf{Method}} 
& \multicolumn{2}{c}{\cellcolor{blue!15}\textbf{Bonn}} 
& \multicolumn{2}{c}{\cellcolor{blue!15}\textbf{TUM Dyn}}
& \multicolumn{2}{c}{\cellcolor{green!15}\textbf{KITTI} }
& \multicolumn{2}{c}{\cellcolor{orange!15}\textbf{PointOdyssey} }
& \multicolumn{2}{c}{\cellcolor{cyan!15}\textbf{Syndrone} }
& \multicolumn{2}{c}{\cellcolor{pink!20}\textbf{Sintel}} \\
& & \cellcolor{blue!15}\scriptsize AbsRel ↓& \cellcolor{blue!15}$\delta{<}1.25$ ↑ 
& \cellcolor{blue!15}\scriptsize AbsRel ↓& \cellcolor{blue!15}$\delta{<}1.25$ ↑
& \cellcolor{green!15}\scriptsize AbsRel ↓& \cellcolor{green!15}$\delta{<}1.25$ ↑
& \cellcolor{orange!15}\scriptsize AbsRel ↓& \cellcolor{orange!15}$\delta{<}1.25$ ↑
& \cellcolor{cyan!15}\scriptsize AbsRel ↓& \cellcolor{cyan!15}$\delta{<}1.25$ ↑
& \cellcolor{pink!20}\scriptsize AbsRel ↓& \cellcolor{pink!20}$\delta{<}1.25$ ↑\\
\midrule
\multirow{15}{*}{\rotatebox{90}{Normalized}}& DepthAnyVideo & 0.515 & 25.3 & 0.184 & 84.6 & 0.074 & 95.3 & 0.417 & 61.7 & 0.299 & 83.1 & 0.455 & 47.9 \\
& VideoDepthAnything & 0.268 & 48.3 & 1.101 & 89.0 & \cellcolor{top2} 0.060 & \cellcolor{top1} 98.2 & 0.283 & 70.3 & 0.138 & 92.5 & 1.691 & 45.4 \\
& DepthCrafter & 0.107 & 88.3 & 0.159 & 79.5 & 0.120 & 86.2 & 0.144 & 81.3 & 0.380 & 87.5 & 0.354 & 58.2 \\
& Marigold & 0.329 & 52.2 & 0.600 & 32.8 & 0.332 & 43.3 & 0.346 & 47.5 & 1.331 & 16.8 & 0.417 & 45.4 \\
\cmidrule(lr){2-14}
& DUSt3R/LSM & 0.174 & 83.5 & 0.187 & 79.2 & 0.124 & 84.9 & 0.168 & 77.8 & \cellcolor{top2} 0.063 & \cellcolor{top2} 96.9 & 0.475 & 59.1 \\
& MASt3R & 0.160 & 81.5 & 0.162 & 83.1 & 0.082 & 93.2 & 0.150 & 79.3 & \cellcolor{top1} 0.046 & \cellcolor{top1} 97.5 & 0.374 & 63.9 \\
& Spann3R & 0.205 & 77.4 & 0.204 & 70.6 & 0.449 & 49.1 & 0.303 & 58.4 & 0.241 & 74.5 & 0.587 & 43.3 \\
& CUT3R & 0.068 & 95.0 & 0.108 & 84.7 & 0.104 & 89.9 & 0.095 & 88.4 & 0.111 & 89.5 & 0.466 & 56.0 \\
& VGGT & \cellcolor{top1} 0.056 & 96.3 & \cellcolor{top1} 0.068 & \cellcolor{top1} 93.9 & \cellcolor{top1} 0.051 & \cellcolor{top2} 96.6 & \cellcolor{top1} 0.026 & \cellcolor{top1} 99.0 & \cellcolor{top3} 0.075 & \cellcolor{top3} 95.9 & \cellcolor{top3} 0.242 & 65.9 \\
& Fast3R & 0.232 & 69.4 & 0.221 & 71.1 & 0.308 & 46.8 & 0.271 & 66.2 & 0.368 & 44.8 & 0.565 & 48.7 \\
& MonST3R & \cellcolor{top3} 0.061 & 95.4 & 0.197 & 72.6 & 0.083 & 93.4 & \cellcolor{top2} 0.066 & 92.3 & 0.110 & 89.7 & 0.343 & 59.4 \\
& Align3R & 0.062 & \cellcolor{top2} 96.8 & \cellcolor{top3} 0.107 & \cellcolor{top3} 90.1 & 0.105 & 89.2 & \cellcolor{top3} 0.077 & \cellcolor{top2} 93.3 & 0.097 & 92.9 & \cellcolor{top2} 0.237 & 69.0 \\
& Easi3R & 0.061 & 95.8 & 0.192 & 76.9 & 0.150 & 76.2 & 0.143 & 82.1 & 0.095 & 94.0 & 0.323 & 53.9 \\
& Geo4D & \cellcolor{top2} 0.060 & \cellcolor{top1} 97.8 & \cellcolor{top2} 0.096 & \cellcolor{top2} 93.2 & 0.086 & 93.8 & 0.082 & \cellcolor{top3} 93.0 & 0.105 & 93.1 & \cellcolor{top1} 0.205 & \cellcolor{top1} 73.2 \\
& Aether & 0.582 & 61.2 & 0.192 & 80.6 & \cellcolor{top3} 0.065 & \cellcolor{top3} 96.2 & 0.123 & 87.9 & 0.145 & 91.1 & 0.343 & \cellcolor{top3} 69.4 \\
& GeometryCrafter & 0.061 & \cellcolor{top3} 96.8 & 0.115 & 87.7 & 0.410 & 53.8 & 0.124 & 83.6 & 0.123 & 90.8 & 0.280 & \cellcolor{top2} 72.4 \\
\cmidrule(lr){1-14}
\multirow{2}{*}{Metric}&MASt3R & 0.549 & 4.6 & 0.633 & 0.9 & 0.754 & 6.4 & 0.749 & 0.2 & 0.967 & 0 & 0.701 & 2.3\\
&CUT3R & 0.097 & 90.3 & 0.135 & 80.6 & 0.118 & 87.4 & 0.127 & 88.1 & 0.824 & 0 & 1.020 & 23.6\\

\bottomrule
\end{tabular}
}
{\tiny
\begin{tabular}{@{}l@{}}
\colorbox{blue!15}{\strut\textbf{Blue}}: Indoor Scene 
\colorbox{green!15}{\strut\textbf{Green}}: Outdoor Scene
\colorbox{orange!15}{\strut\textbf{Orange}}: Large Dynamic Motion
\colorbox{cyan!15}{\strut\textbf{Cyan}}: Drone Scene
\colorbox{pink!20}{\strut\textbf{Pink}}: Mixed Scene
\end{tabular}
}
\vspace{-10pt}
\end{table}

\subsection{Multi-View Relative Pose Estimation}
\label{sec:multi_view_pose}

\textbf{Task:}
\textit{This task evaluates recovering relative camera poses from image collections,} a core capability for downstream tasks like localization, SLAM, and 3D mapping. 
Traditional SfM methods~\cite{schoenberger2016sfm, schoenberger2016mvs} often struggle with sparse views or dynamic motion, 
but recent 3D GFMs show promise in estimating poses directly from raw images without explicit matching. Prior work has typically evaluated static models only on static datasets, and dynamic-aware models on a limited set of dynamic scenes. Our benchmark systematically tests all GFMs across both static and dynamic datasets to assess their generalization to diverse and challenging scenarios. We focus exclusively on relative pose estimation, as most GFMs operate up to an unknown global scale or in relative coordinate frames.

\textbf{Dataset:} We use common datasets: object-centric CO3Dv2~\cite{co3dv2}, indoor static scenes (RealEstate10K~\cite{realestate10k}, ScanNet-eval~\cite{scannet}), indoor dynamic sequences (Bonn~\cite{bonn}, TUM Dynamics~\cite{tumdynamic}), and the indoor-outdoor mixed Sintel~\cite{sintel}. To test generalization, we further include challenging out-of-distribution datasets: KITTI Odometry~\cite{kitti_ordometry} (street driving), ADT~\cite{adt} (egocentric motion with blur), ACID~\cite{acid} and Syndrone~\cite{syndrone} (drone views), and ULTRRA~\cite{ULTRRA} (air-ground paired trajectories with large domain shift). This suite (a total of 11 datasets) covers static and dynamic scenes, indoor and outdoor environments, egocentric and aerial views, across both synthetic and real-world settings.

\textbf{Evaluation Protocol and Metrics:} We report three standard metrics: \textit{Absolute Translation Error} (ATE), \textit{Relative Translation Error} (RPE-trans), and \textit{Relative Rotation Error} (RPE-rot), computed after applying Sim(3) Umeyama alignment~\cite{umeyama1991least} between predicted and ground-truth trajectories. For ULTRRA~\cite{ULTRRA}, where a single Sim(3) alignment is infeasible since the aerial and ground trajectories are reconstructed in separate coordinate systems, we report alternative metrics to ATE in the Appendix~\ref{apx:multi_view_pose_more}.

\textbf{Quantitative Results:} As shown in Tab.~\ref{tab:pose}:
1) Overall, multi-view VGGT and CUT3R, stereo-based methods (DUSt3R, MASt3R with test-time optimization), and diffusion-based Aether rank among the top-performing GFMs.
2) On long video sequences, such as ScanNet-eval, ADT, and TUM Dynamics, GFMs often exhibit increased rotation errors, while driving scenes from KITTI Odometry pose challenges in the form of high ATE.
3) Interestingly, despite being out-of-distribution, drone datasets like ACID and Syndrone are surprisingly well handled, indicating stronger generalization to aerial views than expected.
4) In contrast, GFMs perform significantly worse on the real-world ULTRRA challenge, which involves air-ground input pairs. A specialized method~\cite{vuong2025aerialmegadepth} combining DUSt3R with a diffusion backbone and domain-specific training shows improved performance, with detailed comparisons provided in the Appendix~\ref{apx:multi_view_pose_more}.

\begin{table}[ht]
\centering
\scriptsize
\vspace{-5pt}
\caption{\small \textbf{Evaluation on Multi-view Relative Pose Estimation.} We report ATE (↓), RPE translation (↓), and RPE rotation (↓) across six diverse scene types.  \xmark~indicates methods that are incompatible with pairwise inputs due to requiring a fixed number of input frames.}
\label{tab:pose}
\resizebox{\linewidth}{!}{
\begin{tabular}{lccccccccccccccccc}
\toprule
\multirow{2}{*}{\textbf{Method}} 
& \multicolumn{3}{c}{\cellcolor{gray!20}\textbf{CO3Dv2}} 
& \multicolumn{3}{c}{\cellcolor{blue!15}\textbf{ScanNet \& ADT \& TUM-Dyn.}} 
& \multicolumn{3}{c}{\cellcolor{green!15}\textbf{KITTI Ordometry}} 
& \multicolumn{3}{c}{\cellcolor{orange!15}\textbf{Bonn} \& \textbf{Sintel} \& \textbf{Rel10k}} 
& \multicolumn{3}{c}{\cellcolor{cyan!15}\textbf{ACID \& Syndrone}} 
& \multicolumn{2}{c}{\cellcolor{purple!20}\textbf{ULTRRA}} \\
& \cellcolor{gray!20}ATE & \cellcolor{gray!20}RPE$_\text{trans}$ & \cellcolor{gray!20}RPE$_\text{rot}$
& \cellcolor{blue!15}ATE & \cellcolor{blue!15}RPE$_\text{trans}$ & \cellcolor{blue!15}RPE$_\text{rot}$
& \cellcolor{green!15}ATE & \cellcolor{green!15}RPE$_\text{trans}$ & \cellcolor{green!15}RPE$_\text{rot}$
& \cellcolor{orange!15}ATE & \cellcolor{orange!15}RPE$_\text{trans}$ & 
\cellcolor{orange!15}RPE$_\text{rot}$
& \cellcolor{cyan!15}ATE
& \cellcolor{cyan!15}RPE$_\text{trans}$ & \cellcolor{cyan!15}RPE$_\text{rot}$
& \cellcolor{purple!20}RPE$_\text{trans}$ & \cellcolor{purple!20}RPE$_\text{rot}$ \\
\midrule
DUSt3R/LSM   & 0.903 & 1.325 & \cellcolor{top3}4.312 & \cellcolor{top3}0.139 & \cellcolor{top3}0.102 & \cellcolor{top2}2.394 & 2.935 & 1.135 & 2.832 & 0.077 & 0.557 & 1.657 & 0.126 & 0.379 & 2.836 & 70.350 & 70.390 \\
MASt3R       & 0.987 & 1.407 & \cellcolor{top2}3.999 & \cellcolor{top2}0.131 & \cellcolor{top2}0.098 & 2.889 & \cellcolor{top2}1.492 & \cellcolor{top2}0.399 & \cellcolor{top2}0.407 & \cellcolor{top2}0.058 & 0.559 & 1.305 & 0.130 & 0.376 & 2.601 & 71.519 & 78.036 \\
Spann3R      &0.915 & 1.295 & 6.352 & 0.294 & 0.164 & 3.778 & 15.848 & 5.031 & 4.645 & 0.083 & 0.102 & 1.297 & \cellcolor{top2}0.117 & 0.149 & 1.484 & \cellcolor{top1}40.503 & \cellcolor{top1}38.366 \\
CUT3R        & 0.847 & \cellcolor{top3}1.209 & 6.361 & 0.185 & 0.133 & 4.471 & 2.421 & 0.747 & 0.669 & \cellcolor{top1}0.033 & \cellcolor{top1}0.039 & \cellcolor{top1}0.500 & \cellcolor{top1}0.071 & \cellcolor{top1}0.090 & \cellcolor{top3}0.914 & \cellcolor{top3}55.135 & \cellcolor{top3}54.395 \\
VGGT         & \cellcolor{top1}0.478 & \cellcolor{top1}0.704 & \cellcolor{top1}2.264 & \cellcolor{top1}0.113 & \cellcolor{top1}0.086 & \cellcolor{top1}1.535 & \cellcolor{top1}0.955 & \cellcolor{top1}0.315 & \cellcolor{top1}0.335 & \cellcolor{top3}0.062 & 0.111 & \cellcolor{top2}0.580 & 0.280 & 0.461 & \cellcolor{top2}0.802 & 63.451 & 77.281 \\
Fast3R       & \cellcolor{top2}0.698 & \cellcolor{top2}1.035 & 4.352 & 0.499 & 0.391 & 23.739 & 22.109 & 7.573 & 7.366 & 0.111 & 0.170 & 2.017 & 0.436 & 0.518 & 1.979 & \cellcolor{top2}51.149 & \cellcolor{top2}54.150\\
MonST3R      &2.456 & 3.327 & 23.458 & 0.448 & 0.286 & 12.817 & 2.426 & 0.782 & 0.949 & 0.098 & 0.152 & \cellcolor{top3}0.830 & 0.335 & 0.504 & 1.514 & 70.388 & 77.325\\
Align3R      & 1.027 & 1.550 & 6.499 & 0.425 & 0.215 & 9.430 & 4.611 & 0.817 & \cellcolor{top3}0.600 & 0.076 & \cellcolor{top2}0.091 & 1.083 & 0.150 & 0.179 & 0.977 & 72.010 & 70.638\\
Easi3R       & 0.857 & 1.271 & 5.052 & 0.174 & 0.103 & \cellcolor{top3}2.872 & 3.625 & 0.919 & 0.615 & 0.075 & \cellcolor{top3}0.094 & 1.361 & \cellcolor{top3}0.119 & \cellcolor{top3}0.138 & 1.733 & 62.061 & 71.060 \\
Geo4D        & \cellcolor{top3}0.798 & 1.264 & 5.692 & 0.436 & 0.175 & 10.565 & 1.662 & \cellcolor{top3}0.497 & 0.696 & 0.573 & 0.472 & 3.779 & 0.384 & 0.329 & 1.395 & \xmark & \xmark\\
Aether       &3.168 & 2.366 & 21.643 & 0.644 & 0.273 & 14.804 & \cellcolor{top3}1.553 & 0.744 & 0.744 & 0.195 & 0.122 & 1.610 & 0.152 & \cellcolor{top2}0.097 & \cellcolor{top1}0.796 &\xmark & \xmark \\
\bottomrule
\end{tabular}
}

{\tiny
\resizebox{\linewidth}{!}{
\begin{tabular}{@{}l@{}}
\colorbox{gray!20}{\strut\textbf{Gray}}: In Distribution \quad
\colorbox{blue!15}{\strut\textbf{Blue}}: Long Sequence \quad
\colorbox{green!15}{\strut\textbf{Green}}: Street Driving \quad
\colorbox{orange!15}{\strut\textbf{Orange}}: Indoor-Outdoor Scene \quad
\colorbox{cyan!15}{\strut\textbf{Cyan}}: Drone \quad
\colorbox{purple!20}{\strut\textbf{Purple}}: Air-Ground
\end{tabular}
}
}
\vspace{-10pt}
\end{table}

\subsection{Multi-View 3D Reconstruction}
\label{sec:multi_view_3d}

\textbf{Task:}
\textit{This task assesses reconstructing dense 3D point clouds from multiple input views.} 
This evaluates GFMs for applications such as AR mapping, robot navigation, and neural rendering.
Unlike traditional SfM pipelines like COLMAP~\cite{schoenberger2016sfm, schoenberger2016mvs}, which rely on keypoint matching and multi-stage optimization, end-to-end 3D GFMs can generate pointmaps directly in a feed-forward manner, even under sparse views or dynamic scenes where classical methods often fail. 
Recent studies show that sparse-view predictions from GFMs can be effectively combined with methods like 3D Gaussian Splatting~\cite{fan2024instantsplat} for 3D reconstruction.
However, prior work typically evaluates either sparse-view or dense-view settings alone. 
Our benchmark considers both, providing a more complete picture of each model’s scalability, generalization, and reconstruction fidelity across varied capture conditions.

\textbf{Dataset:} We evaluate on five datasets: the object-centric dataset DTU~\cite{dtu}, and indoor scene datasets including 7-Scenes~\cite{seven_scenes}, Neural RGB-D~\cite{nrgbd}, ScanNet~\cite{scannet}, and TUM RGB-D~\cite{tum_rgbd}, covering both object-scale and scene-scale reconstruction across synthetic and real-world settings.

\textbf{Evaluation Metrics:} \textit{accuracy} (Acc), the mean distance from predicted points to the ground truth, \textit{completeness} (Comp), the mean distance from ground-truth to the predicted surface, and \textit{normal consistency} (NC), the mean cosine similarity between predicted and ground‑truth surface normals.

\textbf{Evaluation Protocol:} We evaluate: (1) \textbf{Extremely Sparse-view reconstruction} simulates real-world constraints with limited inputs, using 2–5 images per scene selected to have minimal or no overlap and wide baselines. (2) \textbf{Dense-view reconstruction} uses 10–50 images per scene, selected to ensure high coverage and significant view overlap.
To ensure fair comparison, for models(e.g., Fast3R~\cite{Yang_2025_Fast3R} and CUT3R~\cite{wang2025cut3r}) that generate both local and global coordinates, we only evaluate the global pointmap to test their direct inference capability. All models are evaluated without access to ground-truth camera parameters during inference. Predicted point clouds are aligned to ground truth using the Umeyama algorithm. Metrics are computed over valid regions, with official masks applied when available.

\textbf{Quantitative Results:} As shown in Tab.~\ref{tab:3d_recon}:
1) Among feed-forward methods, VGGT~\cite{wang2025vggt} is the most competitive overall. In sparse-view settings, CUT3R and FLARE~\cite{zhang2025FLARE} also perform well, particularly on indoor scenes, suggesting they effectively leverage multi-view cues despite limited inputs. 
2) Methods like MonST3R~\cite{zhang2025monstr} and Align3R~\cite{lu2024align3r} underperform in this task; for simplicity, we report only the representative MonST3R among this group.

\begin{table}[ht]
\centering
\scriptsize
\vspace{-5pt}
\caption{\small \textbf{Comparison on Sparse-view and Dense-view 3D Reconstruction.} We report Accuracy (↓), Completeness (↓), and Normal Consistency (↑).}
\label{tab:3d_recon}
\resizebox{\linewidth}{!}{
\begin{tabular}{llccccccccccccccc}
\toprule
\multirow{3}{*}{\textbf{Setting}} & \multirow{3}{*}{\textbf{Method}} 
& \multicolumn{3}{c}{\cellcolor{gray!20}\textbf{DTU}} 
& \multicolumn{3}{c}{\cellcolor{blue!15}\textbf{7-Scenes}} 
& \multicolumn{3}{c}{\cellcolor{blue!15}\textbf{NRGBD}} 
& \multicolumn{3}{c}{\cellcolor{blue!15}\textbf{ScanNet}} 
& \multicolumn{3}{c}{\cellcolor{blue!15}\textbf{TUM-RGBD}} \\
& & \cellcolor{gray!20}ACC ↓ & \cellcolor{gray!20}Comp ↓ & \cellcolor{gray!20}NC ↑ 
& \cellcolor{blue!15}ACC ↓ & \cellcolor{blue!15}Comp ↓ & \cellcolor{blue!15}NC ↑ 
& \cellcolor{blue!15}ACC ↓ & \cellcolor{blue!15}Comp ↓ & \cellcolor{blue!15}NC ↑ 
& \cellcolor{blue!15}ACC ↓ & \cellcolor{blue!15}Comp ↓ & \cellcolor{blue!15}NC ↑ 
& \cellcolor{blue!15}ACC ↓ & \cellcolor{blue!15}Comp ↓ & \cellcolor{blue!15}NC ↑ \\
\midrule
\multirow{8}{*}{\makecell{Extremely \\ Sparse}} & DUSt3R/LSM & \cellcolor{top1}1.731 & \cellcolor{top1}1.936 & \cellcolor{top1}0.786 & 0.146 & 0.181 & 0.744 & 0.144 & 0.154 & 0.867 & 0.474 & 0.420 & 0.714 & 1.108 & 0.746 & 0.724 \\
 & MASt3R     & \cellcolor{top2} 1.895 & \cellcolor{top2} 2.003 & \cellcolor{top2} 0.788 & 0.262 & 0.254 & 0.732 & 0.113 & 0.102 & 0.810 & 0.467 & 0.389 & 0.701 & 0.738 & 0.747 & 0.739\\
 & Spann3R    &6.275 & 5.460 & 0.705 & 0.189 & 0.188 & 0.653 & 0.255 & 0.262 & 0.628 & 0.487 & 0.408 & 0.617 & 1.561 & 1.002 & 0.621
\\
 & FLARE      &3.406 & 3.950 & 0.491 & 0.152 & 0.154 & 0.704 & \cellcolor{top1}0.060 & \cellcolor{top1}0.056 & \cellcolor{top1}0.839 & \cellcolor{top3}0.357 & \cellcolor{top3}0.302 & \cellcolor{top3}0.561 & \cellcolor{top2} 0.515 & \cellcolor{top2} 0.486 & \cellcolor{top2} 0.677
 \\
 & CUT3R      &6.885 & 5.022 & 0.727 & \cellcolor{top2} 0.118 & \cellcolor{top2} 0.142 & \cellcolor{top2} 0.717 & \cellcolor{top3}0.104 & \cellcolor{top3}0.078 & \cellcolor{top3}0.828 & \cellcolor{top2} 0.260 & \cellcolor{top2} 0.238 & \cellcolor{top2} 0.692 & \cellcolor{top3}0.587 & \cellcolor{top3}0.553 & \cellcolor{top3}0.683
\\
 & VGGT       & \cellcolor{top3}2.716 & \cellcolor{top3}2.301 & \cellcolor{top3}0.765 &\cellcolor{top1}0.077 & \cellcolor{top1}0.080 & \cellcolor{top1} 0.762 & \cellcolor{top2} 0.069 & \cellcolor{top2} 0.071 & \cellcolor{top2} 0.903 & \cellcolor{top1} 0.063 & \cellcolor{top1}0.079 &\cellcolor{top1} 0.798 & \cellcolor{top1}0.385 & \cellcolor{top1} 0.331 & \cellcolor{top1}0.747\\
 & Fast3R     & 4.493 & 3.681 & 0.735 & \cellcolor{top3}0.149 &\cellcolor{top3} 0.116 &\cellcolor{top3} 0.692 & 0.361 & 0.201 & 0.782 & 0.546 & 0.306 & 0.621 & 0.955 & 0.630 & 0.627\\
 & MonST3R    & 20.145 & 10.322 & 0.603 & 0.276 & 0.277 & 0.677 & 0.471 & 0.458 & 0.659 & 0.623 & 0.541 & 0.594 & 1.688 & 1.031 & 0.670\\
\midrule
\multirow{8}{*}{Dense} & DUSt3R/LSM & \cellcolor{top1}1.284 & \cellcolor{top1}1.349 & \cellcolor{top1}0.720 & \cellcolor{top2} 0.022 & \cellcolor{top2} 0.029 & \cellcolor{top2} 0.709 & \cellcolor{top2} 0.035 & \cellcolor{top2} 0.024 & \cellcolor{top2} 0.838 & \cellcolor{top2} 0.026 & \cellcolor{top2} 0.025 & \cellcolor{top2} 0.784 & \cellcolor{top3} 0.620 & \cellcolor{top3} 0.474 & \cellcolor{top3} 0.718 \\
 & MASt3R     &\cellcolor{top2} 1.374 & \cellcolor{top2} 1.409 & \cellcolor{top2} 0.723 & \cellcolor{top3} 0.025 & \cellcolor{top3} 0.028 & \cellcolor{top3} 0.697 & \cellcolor{top3} 0.043 & \cellcolor{top3} 0.024 & \cellcolor{top3} 0.809 & \cellcolor{top3} 0.035 & \cellcolor{top3} 0.027 & \cellcolor{top3} 0.757 & \cellcolor{top2} 0.209 & \cellcolor{top2} 0.211 & \cellcolor{top2} 0.708 \\
 & Spann3R    &6.505 & 3.110 & 0.668 & 0.176 & 0.087 & 0.599 & 0.343 & 0.073 & 0.661 & 0.262 & 0.118 & 0.606 & 0.635 & 0.930 & 0.662\\
 & CUT3R      &4.710 & 2.413 & 0.699 & 0.025 & 0.028 & 0.665 & 0.076 & 0.029 & 0.782 & 0.042 & 0.030 & 0.693 & 0.740 & 0.595 & 0.665 \\
 &  VGGT     & \cellcolor{top3} 2.103 & \cellcolor{top3} 1.925 & \cellcolor{top3} 0.748 & \cellcolor{top1}0.019 & \cellcolor{top1}0.032 & \cellcolor{top1}0.659 & \cellcolor{top1}0.015 & \cellcolor{top1}0.012 & \cellcolor{top1}0.874 & \cellcolor{top1}0.016 & \cellcolor{top1}0.021 & \cellcolor{top1}0.728 & \cellcolor{top1}0.065 & \cellcolor{top1}0.091 & \cellcolor{top1}0.692\\
 &  Fast3R      & 3.647 & 2.319 & 0.695 & 0.046 & 0.057 & 0.636 & 0.059 & 0.028 & 0.772 & 0.200 & 0.077 & 0.625 & 0.711 & 0.337 & 0.610\\

 & MonST3R    & 14.455 & 7.508 & 0.636 & 0.100 & 0.091 & 0.648 & 0.336 & 0.246 & 0.665 & 0.346 & 0.293 & 0.599 & 1.138 & 0.948 & 0.591\\
\bottomrule
\end{tabular}
}
{\tiny
\begin{tabular}{@{}l@{}}
\colorbox{gray!20}{\strut\textbf{Gray}}: Object-Centric \quad
\colorbox{blue!15}{\strut\textbf{Blue}}: Indoor Scenes
\end{tabular}
}
\vspace{-10pt}
\end{table}

\subsection{Novel View Synthesis}


\textbf{Task:}~
\textit{This task evaluates synthesizing photorealistic novel views from a few input images, requiring both accurate geometry and appearance modeling.} We benchmark the subset of 3D GFMs with explicit appearance modeling, LSM~\cite{fan2024largespatialmodelendtoend}, NoPoSplat~\cite{ye2024no}, Splatt3r~\cite{smart2024splatt3r}, and FLARE~\cite{zhang2025FLARE}. Unlike prior work that mainly focuses on in-domain data, our benchmark tests generalization across diverse cross-domain scenarios.

\textbf{Dataset:} We evaluate on four datasets, DTU~\cite{dtu}, RealEstate10K~\cite{realestate10k}, ScanNet++~\cite{scannet++}, and ACID~\cite{acid}, covering both object- and scene-level, synthetic and real, indoor and aerial settings.

\textbf{Evaluation Protocol and Metrics:} As NoPoSplat does not support multi-view input, we adopt a 2-view input setting for all methods to ensure consistency. Models predict novel views from two source images. We report three standard metrics: \textit{Peak Signal-to-Noise Ratio} (PSNR)~\cite{psnr}, \textit{Structural Similarity Index} (SSIM)~\cite{ssim}, and \textit{Learned Perceptual Image Patch Similarity} (LPIPS)~\cite{lpips}.

\textbf{Quantitative Results:} Since these models are often trained on a small subset of datasets, we observe a noticeable performance gap between in-domain and out-of-domain settings. Due to space constraints, we include full quantitative results and visual comparisons in the Appendix~\ref{apx:nvs_more}.

\subsection{Inference Efficiency}

\noindent\textbf{Task:} 
Inference efficiency determines the viability of 3D GFMs in latency-sensitive or resource-sensitive domains, e.g., robotics, AR/VR, and interactive systems. 
We measure both runtime and peak memory usage: memory footprint indicates whether the model can run entirely on-device or must offload to slower external memory (e.g., off-chip DRAM), which incurs extra latency penalties~\cite{zhu2024apollo}.




\textbf{Evaluation Protocol and Metrics:} 
For fairness, we test all models on the same hardware, a single 80GB NVIDIA A100 GPU. We vary the number of input views from 2 to 128 and report two metrics: \textit{Inference time} per scene (seconds) and \textit{Peak GPU Memory} (GB), averaged over 10 runs.

\begin{figure*}[!htb]
    \centering
    \includegraphics[width=0.95 \linewidth]{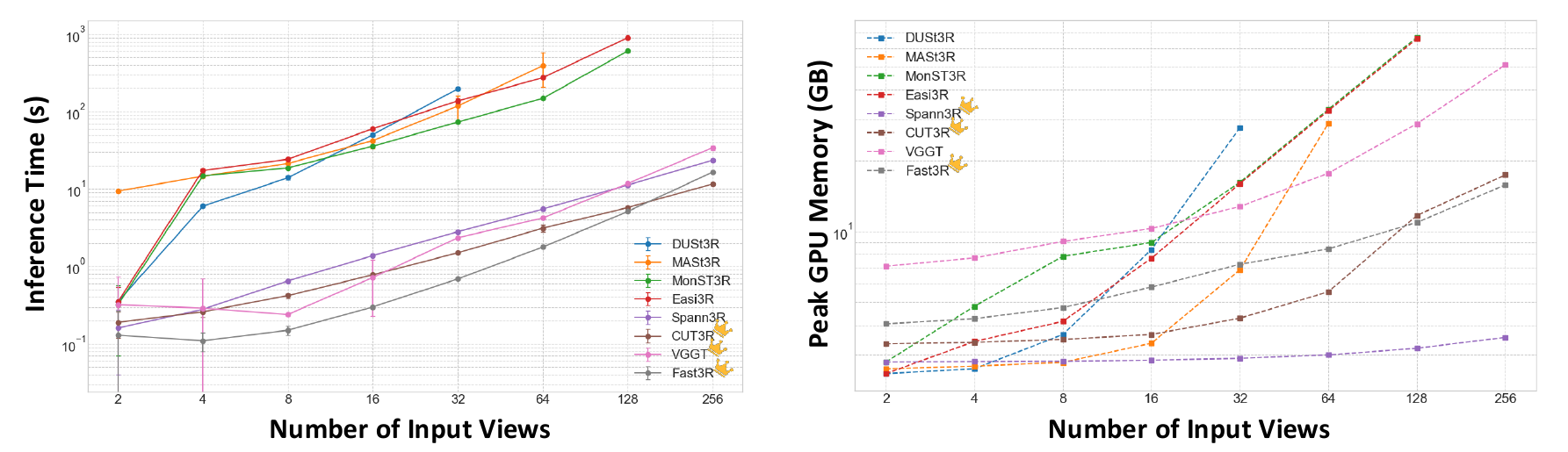}
    \vspace{-2mm}
    \caption{Inference Time (s) (Left) and Peak GPU Memory Usage (GB) (Right) on different numbers of views tested on a single A100. Online registration methods, Spann3R and CUT3R, balance both inference time and GPU consumption.}
    \label{fig:inference_efficiency}
    \vspace{-5pt}
\end{figure*}

\textbf{Quantitative Results:} As shown in
Fig.~\ref{fig:inference_efficiency}):
1) GFMs that need global alignment (GA) incur significantly longer inference times and are more susceptible to out-of-memory (OOM) issues. Techniques such as window-based GA in MonST3R and sparse scene graphs in MASt3R help mitigate these challenges but do not fully resolve them.
2) In contrast, online registration methods like Spann3R~\cite{spann3r} and CUT3R achieve faster inference and lower GPU memory usage, demonstrating better scalability for deployment.

\section{Our Findings}\label{sec:findings}

Beyond simply reporting results and highlighting the top-performing models for each task, we distill and share several high-level insights drawn from our comprehensive evaluation.

\subsection{What Is the Impact of Tasks with Different Difficulties? }\label{sec:rq1}

\textbf{Multi-view geometry inference is inherently harder than pair-view inference.}
Predicting accurate 3D attributes from multiple views requires globally fusing geometric cues, making it a more challenging problem than two‑view reconstruction.
In our experiments, we clearly see the trend that two‑view methods excel in dense‑view settings but fall short under sparse‑view conditions, whereas multi‑view models degrade less(e.g., Tab.~\ref{tab:3d_recon}).
We suspect more data is needed to train the foundational model to handle this harder case, which is confirmed by the fact that the multi-view method VGGT surprisingly performs best on the dense-view setting.
VGGT uses a large dataset with synthetic data, while other underperforming multi-view models usually use relatively smaller datasets.
However, 
although pair‑view approaches can yield higher reconstruction fidelity, their additional alignment steps introduce latency, whereas a single‑pass multi‑view model offers a faster, more streamlined solution.

\textbf{Directly predicting dense 3D scene representations is much more challenging than estimating individual 3D attributes like depth and camera poses.}
Most GFMs achieve strong results on depth estimation, often outperforming task-specific baselines like DepthAnyVideo~\cite{yang2024depthanyvideo}. Their ability to predict complete 3D scenes remains limited, e.g., MonST3R ranks among the top-3 in video depth estimation (Tab.\ref{tab:video_depth}) but performs poorly in 3D reconstruction (Tab.\ref{tab:sparse_depth}), underscoring the increased difficulty of end-to-end global geometry prediction.


\textbf{Metric-scale depth estimation remains a key challenge for GFMs.}
Unlike normalized depth, which is only consistent up to an unknown scale, metric depth requires accurate absolute values—crucial for downstream tasks such as localization, mapping, and planning~\cite{zhou2017unsupervised, yang2020d3vo}. However, learning to predict absolute quantities is inherently more difficult than predicting relative ones. This difficulty is reflected in practice: GFMs that attempt metric-scale estimation still struggle. For example, MASt3R frequently fails across a variety of settings (Tab.~\ref{tab:video_depth} and Tab.~\ref{tab:sparse_depth}), and although CUT3R shows improvements, it performs poorly on drone datasets with large depth ranges and in extreme sparse-view scenarios where spatial priors are hard to establish due to minimal view overlap.


\textbf{Joint prediction of multiple geometric attributes (e.g., pose, depth, matching) may underlie recent performance gains.} Newer GFMs such as CUT3R, Fast3R, and Geo4D jointly predict pointmaps and camera poses, while VGGT further integrates a matching head for keypoint tracking. Such joint learning scheme encourages the network to learn richer and more structured representations that capture both spatial geometry and inter-view relationships, leading to improved generalization and robustness. Similar conclusions were also drawn in the original Geo4D and VGGT works.

\observationbox{
\textit{Takeaway \ding{172}: Current GFMs are promising but face significant challenges when learning from overly complex tasks.}  
\textit{Recommendation:} Carefully decomposing difficult tasks (e.g., jointly predicting geometry, pose, depth, and tracking) into simpler sub-problems can facilitate more effective learning, especially under limited 3D data.
}


\subsection{Do GFMs Generalize Well on Different Data Domains?}\label{sec:rq2}

\textbf{GFMs struggle to generalize in domains with extreme data scarcity.} While GFMs show strong generalization to common out-of-distribution settings, such as aerial views, street-driving scenes, and egocentric perspectives, they often fail in extreme cases like large altitude variation or wide-baseline air-ground pairs, as seen in ULTRRA~\cite{ultrra2}. These failures are largely due to the absence of such examples in existing training data. As demonstrated in \cite{vuong2025aerialmegadepth}, introducing large-scale air-ground paired data substantially improves the performance of DUSt3R variants in these scenarios, confirming that domain generalization can be enhanced through targeted, diverse data. We include additional comparisons and analysis in the Appendix~\ref{apx:multi_view_pose_more}.

This limitation also affects metric-scale depth estimation. Due to the scarcity of metrically accurate training data, GFMs often struggle to predict absolute depth. For instance, CUT3R, trained with broader supervision, outperforms MASt3R across multiple scenarios (Tab.~\ref{tab:video_depth}).


\observationbox{
\textit{Takeaway \ding{173}: Diverse, high-quality data is critical for strong generalization.}
To improve robustness in underrepresented domains, GFMs must be trained on data that covers broader distributions and metric-scale annotations.
}

\subsection{Hints for Model Architecture Design, ViT or Diffusion? Strong 2D Feature Extractor?}\label{sec:rq3}

\textbf{No single design, feed-forward ViT or diffusion, is universally superior.}
While all current GFMs use a transformer-based architecture, they differ in execution: feed-forward ViTs process inputs in a single pass, while diffusion models iteratively refine predictions. Feed-forward designs offer greater flexibility in input configurations and scale more easily with data and modalities, making them well-suited for real-time and general-purpose applications. Diffusion-based GFMs, on the other hand, remain competitive in several tasks (e.g., Tab.~\ref{tab:pose} and Tab.~\ref{tab:video_depth}). These observations suggest that backbone choice should depend on specific task demands and deployment requirements rather than a one-fits-all approach.



\textbf{Stronger 2D foundation models can significantly enhance 3D GFMs.}
Modern 3D GFMs often rely on 2D backbones to extract visual features for 3D reasoning. We observe that more capable 2D feature extractors lead to substantially better performance in 3D tasks. For example, VGGT—despite sharing a similar feed-forward transformer architecture with Fast3R—consistently outperforms it across tasks like sparse/dense 3D reconstruction and video depth estimation. A key factor is VGGT’s use of a DINO-based 2D backbone pretrained on large-scale image datasets, which provides strong visual priors and improves generalization. This highlights the benefit of leveraging powerful 2D foundation models as a stepping stone toward stronger 3D perception.

\observationbox{
\textit{Takeaway \ding{174}: No single backbone—feed‑forward ViT or diffusion, dominates; architecture choice should align with task needs. Moreover, leveraging strong 2D feature extractors (e.g., DINO) substantially boosts 3D performance.}
}

\subsection{Are Current GFMs Ready for Real-Time Perception Systems?}\label{sec:rq4}

\textbf{Despite notable progress, current GFMs are not yet ready for real-time deployment.}
Real-time 3D perception systems require spatial encoders capable of fast and efficient inference. While recent GFMs, especially online registration-based models like Spann3R and CUT3R, show improved latency, none achieve true real-time performance. For example, even the most efficient models still require tens of seconds to process 256 input views (Fig.~\ref{fig:inference_efficiency}), making them unsuitable for time-critical applications such as robotics or AR.

\observationbox{
\textit{Takeaway \ding{175}: As GFMs scale to handle more views and complex tasks, efficiency becomes as critical as accuracy for enabling real-time 3D perception.}
}

\section{Conclusion and Limitations}\label{sec:conclusion}

We present the first comprehensive benchmark evaluating 16 recent end-to-end 3D Geometric Foundation Models (GFMs) across six core tasks, aiming to rigorously assess their effectiveness and efficiency. Our study reveals key insights into the impact of task difficulty, domain generalization, architectural design, and inference efficiency. While GFMs demonstrate strong potential for unified and scalable 3D perception, several challenges remain—particularly in metric-scale depth prediction, sparse-view reconstruction, and real-time deployment.

A limitation of our benchmark is that, in keeping with prior GFM studies, all evaluations use GPU setups. 
Future efforts should explore scaling GFMs through efficient inference techniques such as quantization and sparsity, and extend our evaluations to edge and mobile devices for low‑resource deployment by reusing our benchmarks.



\bibliographystyle{unsrt}
\bibliography{neurips_2025}

\begin{thebibliography}{100}

\bibitem{schoenberger2016mvs}
Johannes~Lutz Sch\"{o}nberger, Enliang Zheng, Marc Pollefeys, and Jan-Michael Frahm.
\newblock Pixelwise view selection for unstructured multi-view stereo.
\newblock In {\em European Conference on Computer Vision (ECCV)}, 2016.

\bibitem{yao2018mvsnet}
Yao Yao, Zixin Luo, Shiwei Li, Tian Fang, and Long Quan.
\newblock Mvsnet: Depth inference for unstructured multi-view stereo.
\newblock In {\em Proceedings of the European conference on computer vision (ECCV)}, pages 767--783, 2018.

\bibitem{gu2020cascade}
Xiaodong Gu, Zhiwen Fan, Siyu Zhu, Zuozhuo Dai, Feitong Tan, and Ping Tan.
\newblock Cascade cost volume for high-resolution multi-view stereo and stereo matching.
\newblock In {\em Proceedings of the IEEE/CVF conference on computer vision and pattern recognition}, pages 2495--2504, 2020.

\bibitem{kerl2013dense}
Christian Kerl, J{\"u}rgen Sturm, and Daniel Cremers.
\newblock Dense visual slam for rgb-d cameras.
\newblock In {\em 2013 IEEE/RSJ international conference on intelligent robots and systems}, pages 2100--2106. IEEE, 2013.

\bibitem{whelan2016elasticfusion}
Thomas Whelan, Renato~F Salas-Moreno, Ben Glocker, Andrew~J Davison, and Stefan Leutenegger.
\newblock Elasticfusion: Real-time dense slam and light source estimation.
\newblock {\em The International Journal of Robotics Research}, 35(14):1697--1716, 2016.

\bibitem{Tateno_2017_CVPR}
Keisuke Tateno, Federico Tombari, Iro Laina, and Nassir Navab.
\newblock Cnn-slam: Real-time dense monocular slam with learned depth prediction.
\newblock In {\em Proceedings of the IEEE Conference on Computer Vision and Pattern Recognition (CVPR)}, July 2017.

\bibitem{Luo-VideoDepth}
Xuan Luo, Jia{-}Bin Huang, Richard Szeliski, Kevin Matzen, and Johannes Kopf.
\newblock Consistent video depth estimation.
\newblock 39(4), 2020.

\bibitem{lasinger2019towards}
Katrin Lasinger, Ren{\'e} Ranftl, Konrad Schindler, and Vladlen Koltun.
\newblock Towards robust monocular depth estimation: Mixing datasets for zero-shot cross-dataset transfer.
\newblock {\em arXiv preprint arXiv:1907.01341}, 2019.

\bibitem{achiam2023gpt}
Josh Achiam, Steven Adler, Sandhini Agarwal, Lama Ahmad, Ilge Akkaya, Florencia~Leoni Aleman, Diogo Almeida, Janko Altenschmidt, Sam Altman, Shyamal Anadkat, et~al.
\newblock Gpt-4 technical report.
\newblock {\em arXiv preprint arXiv:2303.08774}, 2023.

\bibitem{kirillov2023segment}
Alexander Kirillov, Eric Mintun, Nikhila Ravi, Hanzi Mao, Chloe Rolland, Laura Gustafson, Tete Xiao, Spencer Whitehead, Alexander~C Berg, Wan-Yen Lo, et~al.
\newblock Segment anything.
\newblock In {\em Proceedings of the IEEE/CVF international conference on computer vision}, pages 4015--4026, 2023.

\bibitem{pagnoni2024byte}
Artidoro Pagnoni, Ram Pasunuru, Pedro Rodriguez, John Nguyen, Benjamin Muller, Margaret Li, Chunting Zhou, Lili Yu, Jason Weston, Luke Zettlemoyer, et~al.
\newblock Byte latent transformer: Patches scale better than tokens.
\newblock {\em arXiv preprint arXiv:2412.09871}, 2024.

\bibitem{cong2025videolifter}
Wenyan Cong, Hanqing Zhu, Kevin Wang, Jiahui Lei, Colton Stearns, Yuanhao Cai, Dilin Wang, Rakesh Ranjan, Matt Feiszli, Leonidas Guibas, et~al.
\newblock Videolifter: Lifting videos to 3d with fast hierarchical stereo alignment.
\newblock {\em arXiv preprint arXiv:2501.01949}, 2025.

\bibitem{dust3r}
Shuzhe Wang, Vincent Leroy, Yohann Cabon, Boris Chidlovskii, and Jerome Revaud.
\newblock Dust3r: Geometric 3d vision made easy.
\newblock In {\em CVPR}, 2024.

\bibitem{mast3r}
Vincent Leroy, Yohann Cabon, and Jerome Revaud.
\newblock Grounding image matching in 3d with mast3r.
\newblock In {\em ECCV}, 2024.

\bibitem{Yang_2025_Fast3R}
Jianing Yang, Alexander Sax, Kevin~J. Liang, Mikael Henaff, Hao Tang, Ang Cao, Joyce Chai, Franziska Meier, and Matt Feiszli.
\newblock Fast3r: Towards 3d reconstruction of 1000+ images in one forward pass.
\newblock In {\em CVPR}, 2025.

\bibitem{zhang2025monstr}
Junyi Zhang, Charles Herrmann, Junhwa Hur, Varun Jampani, Trevor Darrell, Forrester Cole, Deqing Sun, and Ming-Hsuan Yang.
\newblock Mon{ST}3r: A simple approach for estimating geometry in the presence of motion.
\newblock In {\em ICLR}, 2025.

\bibitem{wang2025vggt}
Jianyuan Wang, Minghao Chen, Nikita Karaev, Andrea Vedaldi, Christian Rupprecht, and David Novotny.
\newblock Vggt: Visual geometry grounded transformer.
\newblock In {\em CVPR}, 2025.

\bibitem{fan2024largespatialmodelendtoend}
Zhiwen Fan, Jian Zhang, Wenyan Cong, Peihao Wang, Renjie Li, Kairun Wen, Shijie Zhou, Achuta Kadambi, Zhangyang Wang, Danfei Xu, Boris Ivanovic, Marco Pavone, and Yue Wang.
\newblock Large spatial model: End-to-end unposed images to semantic 3d.
\newblock In {\em NeurIPS}, 2024.

\bibitem{wang2025cut3r}
Qianqian Wang, Yifei Zhang, Aleksander Holynski, Alexei~A Efros, and Angjoo Kanazawa.
\newblock Continuous 3d perception model with persistent state.
\newblock In {\em CVPR}, 2025.

\bibitem{team2025aether}
Aether Team, Haoyi Zhu, Yifan Wang, Jianjun Zhou, Wenzheng Chang, Yang Zhou, Zizun Li, Junyi Chen, Chunhua Shen, Jiangmiao Pang, et~al.
\newblock Aether: Geometric-aware unified world modeling.
\newblock {\em arXiv preprint arXiv:2503.18945}, 2025.

\bibitem{Geo4D}
Zeren Jiang, Chuanxia Zheng, Iro Laina, Diane Larlus, and Andrea Vedaldi.
\newblock Geo4d: Leveraging video generators for geometric 4d scene reconstruction, 2025.

\bibitem{ye2024no}
Botao Ye, Sifei Liu, Haofei Xu, Xueting Li, Marc Pollefeys, Ming-Hsuan Yang, and Songyou Peng.
\newblock No pose, no problem: Surprisingly simple 3d gaussian splats from sparse unposed images.
\newblock {\em arXiv preprint arXiv:2410.24207}, 2024.

\bibitem{lu2024align3r}
Jiahao Lu, Tianyu Huang, Peng Li, Zhiyang Dou, Cheng Lin, Zhiming Cui, Zhen Dong, Sai-Kit Yeung, Wenping Wang, and Yuan Liu.
\newblock Align3r: Aligned monocular depth estimation for dynamic videos.
\newblock In {\em CVPR}, 2025.

\bibitem{smart2024splatt3r}
Brandon Smart, Chuanxia Zheng, Iro Laina, and Victor~Adrian Prisacariu.
\newblock Splatt3r: Zero-shot gaussian splatting from uncalibrated image pairs, 2024.

\bibitem{chen2025easi3r}
Xingyu Chen, Yue Chen, Yuliang Xiu, Andreas Geiger, and Anpei Chen.
\newblock Easi3r: Estimating disentangled motion from dust3r without training.
\newblock {\em arXiv preprint arXiv:2503.24391}, 2025.

\bibitem{spann3r}
Hengyi Wang and Lourdes Agapito.
\newblock 3d reconstruction with spatial memory.
\newblock In {\em 3DV}, 2025.

\bibitem{xu2025geometrycrafter}
Tian-Xing Xu, Xiangjun Gao, Wenbo Hu, Xiaoyu Li, Song-Hai Zhang, and Ying Shan.
\newblock Geometrycrafter: Consistent geometry estimation for open-world videos with diffusion priors.
\newblock {\em arXiv preprint arXiv:2504.01016}, 2025.

\bibitem{zhang2025FLARE}
Shangzhan Zhang, Jianyuan Wang, Yinghao Xu, Nan Xue, Christian Rupprecht, Xiaowei Zhou, Yujun Shen, and Gordon Wetzstein.
\newblock Flare: Feed-forward geometry, appearance and camera estimation from uncalibrated sparse views.
\newblock In {\em CVPR}, 2025.

\bibitem{dino}
Mathilde Caron, Hugo Touvron, Ishan Misra, Herv\'e J\'egou, Julien Mairal, Piotr Bojanowski, and Armand Joulin.
\newblock Emerging properties in self-supervised vision transformers.
\newblock In {\em Proceedings of the International Conference on Computer Vision (ICCV)}, 2021.

\bibitem{schoenberger2016sfm}
Johannes~Lutz Sch\"{o}nberger and Jan-Michael Frahm.
\newblock Structure-from-motion revisited.
\newblock In {\em Conference on Computer Vision and Pattern Recognition (CVPR)}, 2016.

\bibitem{dtu}
Rasmus Jensen et~al.
\newblock Large scale multi-view stereopsis evaluation.
\newblock In {\em CVPR}, 2014.

\bibitem{eth3d}
Thomas Schops, Johannes~L Schonberger, Silvano Galliani, Torsten Sattler, Konrad Schindler, Marc Pollefeys, and Andreas Geiger.
\newblock A multi-view stereo benchmark with high-resolution images and multi-camera videos.
\newblock In {\em Proceedings of the IEEE conference on computer vision and pattern recognition}, pages 3260--3269, 2017.

\bibitem{kitti}
Andreas Geiger, Philip Lenz, and Raquel Urtasun.
\newblock Are we ready for autonomous driving? the kitti vision benchmark suite.
\newblock In {\em CVPR}, 2012.

\bibitem{tanks}
Arno Knapitsch, Jaesik Park, Qian-Yi Zhou, and Vladlen Koltun.
\newblock Tanks and temples: Benchmarking large-scale scene reconstruction.
\newblock {\em ACM Transactions on Graphics (ToG)}, 36(4):1--13, 2017.

\bibitem{scannet}
Angela Dai, Angel~X. Chang, Manolis Savva, Maciej Halber, Thomas Funkhouser, and Matthias Nie{\ss}ner.
\newblock Scannet: Richly-annotated 3d reconstructions of indoor scenes.
\newblock In {\em CVPR}, 2017.

\bibitem{uhrig2017sparsity}
Jonas Uhrig, Nick Schneider, Lukas Schneider, Uwe Franke, Thomas Brox, and Andreas Geiger.
\newblock Sparsity invariant cnns.
\newblock In {\em 2017 international conference on 3D Vision (3DV)}, pages 11--20. IEEE, 2017.

\bibitem{eigen2014depth}
David Eigen, Christian Puhrsch, and Rob Fergus.
\newblock Depth map prediction from a single image using a multi-scale deep network.
\newblock {\em Advances in neural information processing systems}, 27, 2014.

\bibitem{schroppel2022benchmark}
Max Schroppel, Mathias Rothermel, and et~al.
\newblock A benchmark for multi-view stereo depth estimation under view-point and lighting variations.
\newblock In {\em ECCV}, 2022.

\bibitem{bonn}
Emanuele Palazzolo and Stefan Leutenegger.
\newblock A benchmark for visual-inertial odometry in the presence of motion blur.
\newblock In {\em ICRA}, 2019.

\bibitem{tumdynamic}
J{\"u}rgen Sturm, Nikolas Engelhard, Felix Endres, Wolfram Burgard, and Daniel Cremers.
\newblock A benchmark for the evaluation of rgb-d slam systems.
\newblock In {\em 2012 IEEE/RSJ international conference on intelligent robots and systems}, pages 573--580. IEEE, 2012.

\bibitem{sintel}
DJ~Butler, J~Wulff, GB~Stanley, and MJ~Black.
\newblock A naturalistic open source movie for optical flow evaluation.
\newblock {\em ECCV}, 2012.

\bibitem{pointodyssey}
Boyang Zhao, Renjie Liao, Shiry Yin, et~al.
\newblock Pointodyssey: A large-scale benchmark for robust video depth estimation.
\newblock {\em arXiv preprint arXiv:2402.12345}, 2024.

\bibitem{syndrone}
Giulia Rizzoli, Francesco Barbato, Matteo Caligiuri, and Pietro Zanuttigh.
\newblock Syndrone-multi-modal uav dataset for urban scenarios.
\newblock In {\em Proceedings of the IEEE/CVF International Conference on Computer Vision}, pages 2210--2220, 2023.

\bibitem{video_depth_anything}
Sili Chen, Hengkai Guo, Shengnan Zhu, Feihu Zhang, Zilong Huang, Jiashi Feng, and Bingyi Kang.
\newblock Video depth anything: Consistent depth estimation for super-long videos.
\newblock {\em arXiv:2501.12375}, 2025.

\bibitem{yang2024depthanyvideo}
Honghui Yang, Di~Huang, Wei Yin, Chunhua Shen, Haifeng Liu, Xiaofei He, Binbin Lin, Wanli Ouyang, and Tong He.
\newblock Depth any video with scalable synthetic data.
\newblock {\em arXiv preprint arXiv:2410.10815}, 2024.

\bibitem{ke2025marigold}
Bingxin Ke, Kevin Qu, Tianfu Wang, Nando Metzger, Shengyu Huang, Bo~Li, Anton Obukhov, and Konrad Schindler.
\newblock Marigold: Affordable adaptation of diffusion-based image generators for image analysis, 2025.

\bibitem{hu2024depthcrafter}
Wenbo Hu, Xiangjun Gao, Xiaoyu Li, Sijie Zhao, Xiaodong Cun, Yong Zhang, Long Quan, and Ying Shan.
\newblock Depthcrafter: Generating consistent long depth sequences for open-world videos.
\newblock {\em arXiv preprint arXiv:2409.02095}, 2024.

\bibitem{co3dv2}
Nancy Ruiz, Rohit Varma, Justin Johnson, and Angjoo Kanazawa.
\newblock Learning object-centric representations of multi-object scenes from multiple views.
\newblock In {\em CVPR}, 2022.

\bibitem{realestate10k}
Tinghui Zhou, Matthew Brown, Noah Snavely, and David~G. Lowe.
\newblock Unsupervised learning of depth and ego-motion from video.
\newblock In {\em CVPR}, 2017.

\bibitem{kitti_ordometry}
Andreas Geiger, Philip Lenz, and Raquel Urtasun.
\newblock Are we ready for autonomous driving? the kitti vision benchmark suite.
\newblock In {\em Conference on Computer Vision and Pattern Recognition (CVPR)}, 2012.

\bibitem{adt}
Skanda Koppula, Ignacio Rocco, Yi~Yang, Joe Heyward, Jo{\~a}o Carreira, Andrew Zisserman, Gabriel Brostow, and Carl Doersch.
\newblock Tapvid-3d: A benchmark for tracking any point in 3d.
\newblock {\em arXiv preprint arXiv:2407.05921}, 2024.

\bibitem{acid}
Liyang Zhou et~al.
\newblock Acid: Aerial-captured image dataset for visual localization.
\newblock In {\em ECCV}, 2022.

\bibitem{ULTRRA}
Neil Joshi, Joshua Carney, Nathanael Kuo, Homer Li, Cheng Peng, and Myron Brown.
\newblock Ultrra challenge 2025, 2024.

\bibitem{umeyama1991least}
Shinji Umeyama.
\newblock Least-squares estimation of transformation parameters between two point patterns.
\newblock {\em IEEE Transactions on Pattern Analysis \& Machine Intelligence}, 13(04):376--380, 1991.

\bibitem{vuong2025aerialmegadepth}
Khiem Vuong, Anurag Ghosh, Deva Ramanan, Srinivasa Narasimhan, and Shubham Tulsiani.
\newblock Aerialmegadepth: Learning aerial-ground reconstruction and view synthesis.
\newblock In {\em Proceedings of the IEEE/CVF Conference on Computer Vision and Pattern Recognition}, 2025.

\bibitem{fan2024instantsplat}
Zhiwen Fan, Wenyan Cong, Kairun Wen, Kevin Wang, Jian Zhang, Xinghao Ding, Danfei Xu, Boris Ivanovic, Marco Pavone, Georgios Pavlakos, Zhangyang Wang, and Yue Wang.
\newblock Instantsplat: Sparse-view sfm-free gaussian splatting in seconds, 2024.

\bibitem{seven_scenes}
Jamie Shotton et~al.
\newblock Scene coordinate regression forests for camera relocalization in rgb-d images.
\newblock In {\em CVPR}, 2013.

\bibitem{nrgbd}
Guandao Yang et~al.
\newblock Nrgbd: A large-scale dataset for novel view synthesis and 3d reconstruction from rgb-d images.
\newblock In {\em NeurIPS Datasets and Benchmarks}, 2023.

\bibitem{tum_rgbd}
J{"u}rgen Sturm, Nikolas Engelhard, Felix Endres, Wolfram Burgard, and Daniel Cremers.
\newblock A benchmark for the evaluation of rgb-d slam systems.
\newblock {\em IROS}, 2012.

\bibitem{scannet++}
Chandan Yeshwanth, Yueh-Cheng Liu, Matthias Nie{\ss}ner, and Angela Dai.
\newblock Scannet++: A high-fidelity dataset of 3d indoor scenes.
\newblock In {\em Proceedings of the IEEE/CVF International Conference on Computer Vision}, pages 12--22, 2023.

\bibitem{psnr}
Aljoscha Hore and Djemel Ziou.
\newblock Image quality metrics: Psnr vs. ssim.
\newblock {\em Pattern Recognition}, 2010.

\bibitem{ssim}
Zhou Wang, Alan~C. Bovik, Hamid~R. Sheikh, and Eero~P. Simoncelli.
\newblock Image quality assessment: From error visibility to structural similarity.
\newblock {\em IEEE Transactions on Image Processing}, 2004.

\bibitem{lpips}
Richard Zhang, Phillip Isola, Alexei~A. Efros, Eli Shechtman, and Oliver Wang.
\newblock The unreasonable effectiveness of deep features as a perceptual metric.
\newblock In {\em CVPR}, 2018.

\bibitem{zhu2024apollo}
Hanqing Zhu, Zhenyu Zhang, Wenyan Cong, Xi~Liu, Sem Park, Vikas Chandra, Bo~Long, David~Z Pan, Zhangyang Wang, and Jinwon Lee.
\newblock Apollo: Sgd-like memory, adamw-level performance.
\newblock {\em arXiv preprint arXiv:2412.05270}, 2024.

\bibitem{zhou2017unsupervised}
Tinghui Zhou and et~al.
\newblock Unsupervised learning of depth and ego-motion from video.
\newblock In {\em CVPR}, 2017.

\bibitem{yang2020d3vo}
Nan Yang, Rui Wang, and Daniel Cremers.
\newblock D3vo: Deep depth, deep pose and deep uncertainty for monocular visual odometry.
\newblock In {\em CVPR}, 2020.

\bibitem{ultrra2}
Zihan Wang et~al.
\newblock Ultrra: A benchmark for air-ground relative pose estimation.
\newblock In {\em CVPR}, 2024.

\bibitem{lu2024lora3d}
Ziqi Lu, Heng Yang, Danfei Xu, Boyi Li, Boris Ivanovic, Marco Pavone, and Yue Wang.
\newblock Lora3d: Low-rank self-calibration of 3d geometric foundation models.
\newblock {\em arXiv preprint arXiv:2412.07746}, 2024.

\bibitem{croco}
{Weinzaepfel, Philippe and Leroy, Vincent and Lucas, Thomas and Br\'egier, Romain and Cabon, Yohann and Arora, Vaibhav and Antsfeld, Leonid and Chidlovskii, Boris and Csurka, Gabriela and Revaud J\'er\^ome}.
\newblock {CroCo: Self-Supervised Pre-training for 3D Vision Tasks by Cross-View Completion}.
\newblock In {\em NeurIPS}, 2022.

\bibitem{croco_v2}
Philippe Weinzaepfel, Thomas Lucas, Vincent Leroy, Yohann Cabon, Vaibhav Arora, Romain Br{\'e}gier, Gabriela Csurka, Leonid Antsfeld, Boris Chidlovskii, and J{\'e}r{\^o}me Revaud.
\newblock {CroCo v2: Improved Cross-view Completion Pre-training for Stereo Matching and Optical Flow}.
\newblock In {\em ICCV}, 2023.

\bibitem{jin2024stereo4d}
Linyi Jin, Richard Tucker, Zhengqi Li, David Fouhey, Noah Snavely, and Aleksander Holynski.
\newblock Stereo4d: Learning how things move in 3d from internet stereo videos.
\newblock In {\em CVPR}, 2025.

\bibitem{yao2025uni4d}
David~Yifan Yao, Albert~J Zhai, and Shenlong Wang.
\newblock Uni4d: Unifying visual foundation models for 4d modeling from a single video.
\newblock In {\em Proceedings of the Computer Vision and Pattern Recognition Conference}, pages 1116--1126, 2025.

\bibitem{Liang2025ZeroShotMSF}
Yiqing Liang, Abhishek Badki, Hang Su, James Tompkin, and Orazio Gallo.
\newblock Zero-shot monocular scene flow estimation in the wild.
\newblock In {\em CVPR}, 2025.

\bibitem{sucar2025dynamic}
Edgar Sucar, Zihang Lai, Eldar Insafutdinov, and Andrea Vedaldi.
\newblock Dynamic point maps: A versatile representation for dynamic 3d reconstruction.
\newblock {\em arXiv preprint arXiv:2503.16318}, 2025.

\bibitem{han2025d}
Jisang Han, Honggyu An, Jaewoo Jung, Takuya Narihira, Junyoung Seo, Kazumi Fukuda, Chaehyun Kim, Sunghwan Hong, Yuki Mitsufuji, and Seungryong Kim.
\newblock D\^{} 2ust3r: Enhancing 3d reconstruction with 4d pointmaps for dynamic scenes.
\newblock {\em arXiv preprint arXiv:2504.06264}, 2025.

\bibitem{park2025learning}
Seong~Hyeon Park and Jinwoo Shin.
\newblock Learning multi-frame and monocular prior for estimating geometry in dynamic scenes.
\newblock {\em arXiv preprint arXiv:2505.01737}, 2025.

\bibitem{zhang2025pomato}
Songyan Zhang, Yongtao Ge, Jinyuan Tian, Guangkai Xu, Hao Chen, Chen Lv, and Chunhua Shen.
\newblock Pomato: Marrying pointmap matching with temporal motion for dynamic 3d reconstruction.
\newblock {\em arXiv preprint arXiv:2504.05692}, 2025.

\bibitem{chen2025back}
Weirong Chen, Ganlin Zhang, Felix Wimbauer, Rui Wang, Nikita Araslanov, Andrea Vedaldi, and Daniel Cremers.
\newblock Back on track: Bundle adjustment for dynamic scene reconstruction.
\newblock {\em arXiv preprint arXiv:2504.14516}, 2025.

\bibitem{yang2024storm}
Jiawei Yang, Jiahui Huang, Yuxiao Chen, Yan Wang, Boyi Li, Yurong You, Apoorva Sharma, Maximilian Igl, Peter Karkus, Danfei Xu, et~al.
\newblock Storm: Spatio-temporal reconstruction model for large-scale outdoor scenes.
\newblock {\em arXiv preprint arXiv:2501.00602}, 2024.

\bibitem{sucar2025dynamicpointmapsversatile}
Edgar Sucar, Zihang Lai, Eldar Insafutdinov, and Andrea Vedaldi.
\newblock Dynamic point maps: A versatile representation for dynamic 3d reconstruction, 2025.

\bibitem{liu2025regist3r}
Sidun Liu, Wenyu Li, Peng Qiao, and Yong Dou.
\newblock Regist3r: Incremental registration with stereo foundation model.
\newblock {\em arXiv preprint arXiv:2504.12356}, 2025.

\bibitem{duisterhof2024mast3r}
Bardienus Duisterhof, Lojze Zust, Philippe Weinzaepfel, Vincent Leroy, Yohann Cabon, and Jerome Revaud.
\newblock Mast3r-sfm: a fully-integrated solution for unconstrained structure-from-motion.
\newblock {\em arXiv preprint arXiv:2409.19152}, 2024.

\bibitem{must3r}
Yohann Cabon, Lucas Stoffl, Leonid Antsfeld, Gabriela Csurka, Boris Chidlovskii, Jerome Revaud, and Vincent Leroy.
\newblock Must3r: Multi-view network for stereo 3d reconstruction.
\newblock In {\em Proceedings of the Computer Vision and Pattern Recognition Conference (CVPR)}, pages 1050--1060, June 2025.

\bibitem{light3r_sfm}
Sven Elflein, Qunjie Zhou, and Laura Leal-Taix\'e.
\newblock Light3r-sfm: Towards feed-forward structure-from-motion.
\newblock In {\em Proceedings of the Computer Vision and Pattern Recognition Conference (CVPR)}, pages 16774--16784, June 2025.

\bibitem{tang2025mv}
Zhenggang Tang, Yuchen Fan, Dilin Wang, Hongyu Xu, Rakesh Ranjan, Alexander Schwing, and Zhicheng Yan.
\newblock Mv-dust3r+: Single-stage scene reconstruction from sparse views in 2 seconds.
\newblock In {\em Proceedings of the Computer Vision and Pattern Recognition Conference}, pages 5283--5293, 2025.

\bibitem{sheng2025spatialsplat}
Yu~Sheng, Jiajun Deng, Xinran Zhang, Yu~Zhang, Bei Hua, Yanyong Zhang, and Jianmin Ji.
\newblock Spatialsplat: Efficient semantic 3d from sparse unposed images.
\newblock {\em arXiv preprint arXiv:2505.23044}, 2025.

\bibitem{lu2025matrix3d}
Yuanxun Lu, Jingyang Zhang, Tian Fang, Jean-Daniel Nahmias, Yanghai Tsin, Long Quan, Xun Cao, Yao Yao, and Shiwei Li.
\newblock Matrix3d: Large photogrammetry model all-in-one.
\newblock {\em arXiv preprint arXiv:2502.07685}, 2025.

\bibitem{jiang2025anysplat}
Lihan Jiang, Yucheng Mao, Linning Xu, Tao Lu, Kerui Ren, Yichen Jin, Xudong Xu, Mulin Yu, Jiangmiao Pang, Feng Zhao, et~al.
\newblock Anysplat: Feed-forward 3d gaussian splatting from unconstrained views.
\newblock {\em arXiv preprint arXiv:2505.23716}, 2025.

\bibitem{li2025vicasplat}
Zhiqi Li, Chengrui Dong, Yiming Chen, Zhangchi Huang, and Peidong Liu.
\newblock Vicasplat: A single run is all you need for 3d gaussian splatting and camera estimation from unposed video frames.
\newblock {\em arXiv preprint arXiv:2503.10286}, 2025.

\bibitem{chen2024pref3r}
Zequn Chen, Jiezhi Yang, and Heng Yang.
\newblock Pref3r: Pose-free feed-forward 3d gaussian splatting from variable-length image sequence.
\newblock {\em arXiv preprint arXiv:2411.16877}, 2024.

\bibitem{jang2025pow3r}
Wonbong Jang, Philippe Weinzaepfel, Vincent Leroy, Lourdes Agapito, and Jerome Revaud.
\newblock Pow3r: Empowering unconstrained 3d reconstruction with camera and scene priors.
\newblock {\em arXiv preprint arXiv:2503.17316}, 2025.

\bibitem{raj2024spurfiessparsesurfacereconstruction}
Kevin Raj, Christopher Wewer, Raza Yunus, Eddy Ilg, and Jan~Eric Lenssen.
\newblock Spurfies: Sparse surface reconstruction using local geometry priors, 2024.

\bibitem{liu2025towards}
Jiachen Liu, Rui Yu, Sili Chen, Sharon~X Huang, and Hengkai Guo.
\newblock Towards in-the-wild 3d plane reconstruction from a single image.
\newblock In {\em Proceedings of the Computer Vision and Pattern Recognition Conference}, pages 27027--27037, 2025.

\bibitem{li2024_MegaSaM}
Zhengqi Li, Richard Tucker, Forrester Cole, Qianqian Wang, Linyi Jin, Vickie Ye, Angjoo Kanazawa, Aleksander Holynski, and Noah Snavely.
\newblock {MegaSaM}: Accurate, fast and robust structure and motion from casual dynamic videos.
\newblock In {\em CVPR}, 2025.

\bibitem{vggtslam}
Dominic Maggio, Hyungtae Lim, and Luca Carlone.
\newblock Vggt-slam: Dense rgb slam optimized on the sl (4) manifold.
\newblock {\em arXiv preprint arXiv:2505.12549}, 2025.

\bibitem{mast3r_slam}
Riku Murai, Eric Dexheimer, and Andrew~J. Davison.
\newblock Mast3r-slam: Real-time dense slam with 3d reconstruction priors.
\newblock In {\em Proceedings of the Computer Vision and Pattern Recognition Conference (CVPR)}, pages 16695--16705, June 2025.

\bibitem{wildgs_slam}
Jianhao Zheng, Zihan Zhu, Valentin Bieri, Marc Pollefeys, Songyou Peng, and Iro Armeni.
\newblock Wildgs-slam: Monocular gaussian splatting slam in dynamic environments.
\newblock In {\em Proceedings of the Computer Vision and Pattern Recognition Conference (CVPR)}, pages 11461--11471, June 2025.

\bibitem{slam3r}
Yuzheng Liu, Siyan Dong, Shuzhe Wang, Yingda Yin, Yanchao Yang, Qingnan Fan, and Baoquan Chen.
\newblock Slam3r: Real-time dense scene reconstruction from monocular rgb videos.
\newblock In {\em Proceedings of the Computer Vision and Pattern Recognition Conference (CVPR)}, pages 16651--16662, June 2025.

\bibitem{li2025hier}
Boying Li, Vuong~Chi Hao, Peter~J Stuckey, Ian Reid, and Hamid Rezatofighi.
\newblock Hier-slam++: Neuro-symbolic semantic slam with a hierarchically categorical gaussian splatting.
\newblock {\em arXiv preprint arXiv:2502.14931}, 2025.

\bibitem{fei2024driv3r}
Xin Fei, Wenzhao Zheng, Yueqi Duan, Wei Zhan, Masayoshi Tomizuka, Kurt Keutzer, and Jiwen Lu.
\newblock Driv3r: Learning dense 4d reconstruction for autonomous driving.
\newblock {\em arXiv preprint arXiv:2412.06777}, 2024.

\bibitem{li2025rig3r}
Samuel Li, Pujith Kachana, Prajwal Chidananda, Saurabh Nair, Yasutaka Furukawa, and Matthew Brown.
\newblock Rig3r: Rig-aware conditioning for learned 3d reconstruction.
\newblock {\em arXiv preprint arXiv:2506.02265}, 2025.

\bibitem{muller2025reconstructing}
Lea M{\"u}ller, Hongsuk Choi, Anthony Zhang, Brent Yi, Jitendra Malik, and Angjoo Kanazawa.
\newblock Reconstructing people, places, and cameras.
\newblock In {\em Proceedings of the Computer Vision and Pattern Recognition Conference}, pages 21948--21958, 2025.

\bibitem{liu2025joint}
Zhizheng Liu, Joe Lin, Wayne Wu, and Bolei Zhou.
\newblock Joint optimization for 4d human-scene reconstruction in the wild.
\newblock {\em arXiv preprint arXiv:2501.02158}, 2025.

\bibitem{fan2025vlm3r}
Zhiwen Fan, Jian Zhang, Renjie Li, Junge Zhang, Runjin Chen, Hezhen Hu, Kevin Wang, Huaizhi Qu, Dilin Wang, Zhicheng Yan, Hongyu Xu, Justin Theiss, Tianlong Chen, Jiachen Li, Zhengzhong Tu, Zhangyang Wang, and Rakesh Ranjan.
\newblock Vlm-3r: Vision-language models augmented with instruction-aligned 3d reconstruction, 2025.

\bibitem{diffusionsfm}
Qitao Zhao, Amy Lin, Jeff Tan, Jason~Y. Zhang, Deva Ramanan, and Shubham Tulsiani.
\newblock Diffusionsfm: Predicting structure and motion via ray origin and endpoint diffusion.
\newblock In {\em Proceedings of the Computer Vision and Pattern Recognition Conference (CVPR)}, pages 6317--6326, June 2025.

\bibitem{sun2025unigeo}
Yang-Tian Sun, Xin Yu, Zehuan Huang, Yi-Hua Huang, Yuan-Chen Guo, Ziyi Yang, Yan-Pei Cao, and Xiaojuan Qi.
\newblock Unigeo: Taming video diffusion for unified consistent geometry estimation.
\newblock {\em arXiv preprint arXiv:2505.24521}, 2025.

\bibitem{wang2024vggsfm}
Jianyuan Wang, Nikita Karaev, Christian Rupprecht, and David Novotny.
\newblock Vggsfm: Visual geometry grounded deep structure from motion.
\newblock In {\em Proceedings of the IEEE/CVF conference on computer vision and pattern recognition}, pages 21686--21697, 2024.

\bibitem{smith2024flowmap}
Cameron Smith, David Charatan, Ayush Tewari, and Vincent Sitzmann.
\newblock Flowmap: High-quality camera poses, intrinsics, and depth via gradient descent.
\newblock {\em arXiv preprint arXiv:2404.15259}, 2024.

\bibitem{ranftl2020towards}
Ren{\'e} Ranftl, Katrin Lasinger, David Hafner, Konrad Schindler, and Vladlen Koltun.
\newblock Towards robust monocular depth estimation: Mixing datasets for zero-shot cross-dataset transfer.
\newblock {\em IEEE transactions on pattern analysis and machine intelligence}, 44(3):1623--1637, 2020.

\bibitem{yang2024depthanything}
Lihe Yang, Bingyi Kang, Zilong Huang, Xiaogang Xu, Jiashi Feng, and Hengshuang Zhao.
\newblock Depth anything: Unleashing the power of large-scale unlabeled data.
\newblock In {\em Proceedings of the IEEE/CVF Conference on Computer Vision and Pattern Recognition}, pages 10371--10381, 2024.

\bibitem{depth_anything_v2}
Lihe Yang, Bingyi Kang, Zilong Huang, Zhen Zhao, Xiaogang Xu, Jiashi Feng, and Hengshuang Zhao.
\newblock Depth anything v2.
\newblock {\em arXiv:2406.09414}, 2024.

\bibitem{ke2024repurposing}
Bingxin Ke, Anton Obukhov, Shengyu Huang, Nando Metzger, Rodrigo~Caye Daudt, and Konrad Schindler.
\newblock Repurposing diffusion-based image generators for monocular depth estimation.
\newblock In {\em Proceedings of the IEEE/CVF Conference on Computer Vision and Pattern Recognition}, pages 9492--9502, 2024.

\bibitem{garcia2025fine}
Gonzalo~Martin Garcia, Karim Abou~Zeid, Christian Schmidt, Daan De~Geus, Alexander Hermans, and Bastian Leibe.
\newblock Fine-tuning image-conditional diffusion models is easier than you think.
\newblock In {\em 2025 IEEE/CVF Winter Conference on Applications of Computer Vision (WACV)}, pages 753--762. IEEE, 2025.

\bibitem{fu2024geowizard}
Xiao Fu, Wei Yin, Mu~Hu, Kaixuan Wang, Yuexin Ma, Ping Tan, Shaojie Shen, Dahua Lin, and Xiaoxiao Long.
\newblock Geowizard: Unleashing the diffusion priors for 3d geometry estimation from a single image.
\newblock In {\em European Conference on Computer Vision}, pages 241--258. Springer, 2024.

\bibitem{gui2024depthfm}
Ming Gui, Johannes Schusterbauer, Ulrich Prestel, Pingchuan Ma, Dmytro Kotovenko, Olga Grebenkova, Stefan~Andreas Baumann, Vincent~Tao Hu, and Bj{\"o}rn Ommer.
\newblock Depthfm: Fast monocular depth estimation with flow matching.
\newblock {\em arXiv preprint arXiv:2403.13788}, 2024.

\bibitem{he2024lotus}
Jing He, Haodong Li, Wei Yin, Yixun Liang, Leheng Li, Kaiqiang Zhou, Hongbo Zhang, Bingbing Liu, and Ying-Cong Chen.
\newblock Lotus: Diffusion-based visual foundation model for high-quality dense prediction.
\newblock {\em arXiv preprint arXiv:2409.18124}, 2024.

\bibitem{pham2024sharpdepth}
Duc-Hai Pham, Tung Do, Phong Nguyen, Binh-Son Hua, Khoi Nguyen, and Rang Nguyen.
\newblock Sharpdepth: Sharpening metric depth predictions using diffusion distillation.
\newblock {\em arXiv preprint arXiv:2411.18229}, 2024.

\bibitem{zhang2021consistent}
Zhoutong Zhang, Forrester Cole, Richard Tucker, William~T Freeman, and Tali Dekel.
\newblock Consistent depth of moving objects in video.
\newblock {\em ACM Transactions on Graphics (ToG)}, 40(4):1--12, 2021.

\bibitem{yasarla2023mamo}
Rajeev Yasarla, Hong Cai, Jisoo Jeong, Yunxiao Shi, Risheek Garrepalli, and Fatih Porikli.
\newblock Mamo: Leveraging memory and attention for monocular video depth estimation.
\newblock In {\em Proceedings of the IEEE/CVF International Conference on Computer Vision}, pages 8754--8764, 2023.

\bibitem{wang2023neural}
Yiran Wang, Min Shi, Jiaqi Li, Zihao Huang, Zhiguo Cao, Jianming Zhang, Ke~Xian, and Guosheng Lin.
\newblock Neural video depth stabilizer.
\newblock In {\em Proceedings of the IEEE/CVF International Conference on Computer Vision}, pages 9466--9476, 2023.

\bibitem{kopf2021robust}
Johannes Kopf, Xuejian Rong, and Jia-Bin Huang.
\newblock Robust consistent video depth estimation.
\newblock In {\em Proceedings of the IEEE/CVF Conference on Computer Vision and Pattern Recognition}, pages 1611--1621, 2021.

\bibitem{luo2020consistent}
Xuan Luo, Jia-Bin Huang, Richard Szeliski, Kevin Matzen, and Johannes Kopf.
\newblock Consistent video depth estimation.
\newblock {\em ACM Transactions on Graphics (ToG)}, 39(4):71--1, 2020.

\bibitem{shao2024learning}
Jiahao Shao, Yuanbo Yang, Hongyu Zhou, Youmin Zhang, Yujun Shen, Vitor Guizilini, Yue Wang, Matteo Poggi, and Yiyi Liao.
\newblock Learning temporally consistent video depth from video diffusion priors.
\newblock {\em arXiv preprint arXiv:2406.01493}, 2024.

\bibitem{yin2021learning}
Wei Yin, Jianming Zhang, Oliver Wang, Simon Niklaus, Long Mai, Simon Chen, and Chunhua Shen.
\newblock Learning to recover 3d scene shape from a single image.
\newblock In {\em Proceedings of the IEEE/CVF Conference on Computer Vision and Pattern Recognition}, pages 204--213, 2021.

\bibitem{yin2022towards}
Wei Yin, Jianming Zhang, Oliver Wang, Simon Niklaus, Simon Chen, Yifan Liu, and Chunhua Shen.
\newblock Towards accurate reconstruction of 3d scene shape from a single monocular image.
\newblock {\em IEEE Transactions on Pattern Analysis and Machine Intelligence}, 45(5):6480--6494, 2022.

\bibitem{piccinelli2024unidepth}
Luigi Piccinelli, Yung-Hsu Yang, Christos Sakaridis, Mattia Segu, Siyuan Li, Luc Van~Gool, and Fisher Yu.
\newblock Unidepth: Universal monocular metric depth estimation.
\newblock In {\em Proceedings of the IEEE/CVF Conference on Computer Vision and Pattern Recognition}, pages 10106--10116, 2024.

\bibitem{bochkovskii2024depth}
Aleksei Bochkovskii, Ama{\~A}{\c{G}}l Delaunoy, Hugo Germain, Marcel Santos, Yichao Zhou, Stephan~R Richter, and Vladlen Koltun.
\newblock Depth pro: Sharp monocular metric depth in less than a second.
\newblock {\em arXiv preprint arXiv:2410.02073}, 2024.

\bibitem{wang2024moge}
Ruicheng Wang, Sicheng Xu, Cassie Dai, Jianfeng Xiang, Yu~Deng, Xin Tong, and Jiaolong Yang.
\newblock Moge: Unlocking accurate monocular geometry estimation for open-domain images with optimal training supervision.
\newblock {\em arXiv preprint arXiv:2410.19115}, 2024.

\bibitem{yin2023metric3d}
Wei Yin, Chi Zhang, Hao Chen, Zhipeng Cai, Gang Yu, Kaixuan Wang, Xiaozhi Chen, and Chunhua Shen.
\newblock Metric3d: Towards zero-shot metric 3d prediction from a single image.
\newblock In {\em Proceedings of the IEEE/CVF International Conference on Computer Vision}, pages 9043--9053, 2023.

\bibitem{hu2024metric3d}
Mu~Hu, Wei Yin, Chi Zhang, Zhipeng Cai, Xiaoxiao Long, Hao Chen, Kaixuan Wang, Gang Yu, Chunhua Shen, and Shaojie Shen.
\newblock Metric3d v2: A versatile monocular geometric foundation model for zero-shot metric depth and surface normal estimation.
\newblock {\em IEEE Transactions on Pattern Analysis and Machine Intelligence}, 2024.

\bibitem{charatan2023pixelsplat}
David Charatan, Sizhe~Lester Li, Andrea Tagliasacchi, and Vincent Sitzmann.
\newblock Pixelsplat: 3d gaussian splats from image pairs for scalable generalizable 3d reconstruction. 2024 ieee.
\newblock In {\em CVF Conference on Computer Vision and Pattern Recognition (CVPR)}, pages 19457--19467, 2023.

\bibitem{chen2024mvsplat}
Yuedong Chen, Haofei Xu, Chuanxia Zheng, Bohan Zhuang, Marc Pollefeys, Andreas Geiger, Tat-Jen Cham, and Jianfei Cai.
\newblock Mvsplat: Efficient 3d gaussian splatting from sparse multi-view images.
\newblock In {\em European Conference on Computer Vision}, pages 370--386. Springer, 2024.

\bibitem{khazatsky2024droid}
Alexander Khazatsky, Karl Pertsch, Suraj Nair, Ashwin Balakrishna, Sudeep Dasari, Siddharth Karamcheti, Soroush Nasiriany, Mohan~Kumar Srirama, Lawrence~Yunliang Chen, Kirsty Ellis, et~al.
\newblock Droid: A large-scale in-the-wild robot manipulation dataset.
\newblock {\em arXiv preprint arXiv:2403.12945}, 2024.

\bibitem{li2025sekai}
Zhen Li, Chuanhao Li, Xiaofeng Mao, Shaoheng Lin, Ming Li, Shitian Zhao, Zhaopan Xu, Xinyue Li, Yukang Feng, Jianwen Sun, et~al.
\newblock Sekai: A video dataset towards world exploration.
\newblock {\em arXiv preprint arXiv:2506.15675}, 2025.

\bibitem{wu2025spatial}
Diankun Wu, Fangfu Liu, Yi-Hsin Hung, and Yueqi Duan.
\newblock Spatial-mllm: Boosting mllm capabilities in visual-based spatial intelligence.
\newblock {\em arXiv preprint arXiv:2505.23747}, 2025.

\bibitem{huang2025mllms}
Xiaohu Huang, Jingjing Wu, Qunyi Xie, and Kai Han.
\newblock Mllms need 3d-aware representation supervision for scene understanding.
\newblock {\em arXiv preprint arXiv:2506.01946}, 2025.

\bibitem{caothousands}
Ang Cao, Sergio Arnaud, Oleksandr Maksymets, Jianing Yang, Ayush Jain, Ada Martin, Vincent-Pierre Berges, Paul McVay, Ruslan Partsey, Aravind Rajeswaran, et~al.
\newblock From thousands to billions: 3d visual language grounding via render-supervised distillation from 2d vlms.
\newblock In {\em Forty-second International Conference on Machine Learning}.

\end{thebibliography}

\newpage
\renewcommand{\thepage}{S\arabic{page}}
\renewcommand{\thetable}{S\arabic{table}}   
\renewcommand{\thefigure}{S\arabic{figure}}
\setcounter{figure}{0} 
\setcounter{table}{0} 

\appendix

\section{More Related Works}
\label{apx:related_work}
\subsection{End-to-End 3D Reconstruction}
In this work, we define end-to-end 3D reconstruction as the task of \textit{using fully differentiable models to directly map raw pixels from multiple images into pixel-aligned 3D point maps, without requiring explicit camera intrinsics or extrinsics at inference time.}
End-to-end 3D geometric foundation models (GFMs)\cite{lu2024lora3d}, which enable dense 3D perception directly from raw images, have recently emerged as powerful alternatives to traditional multi-stage pipelines\cite{schoenberger2016sfm, schoenberger2016mvs}.
Pioneering work such as DUSt3R~\cite{dust3r}, pretrained on large-scale image correspondences~\cite{croco,croco_v2}, predicts dense pointmaps from stereo pairs and implicitly estimates correspondences—achieving state-of-the-art performance in depth and pose estimation.

\textbf{Stream 1: Feed-forward GFMs extending DUSt3R.}
A major branch of GFMs evolves from DUSt3R, extending it across multiple axes, including metric-scale prediction, dynamic scene modeling, and multi-view inference. For instance, MASt3R~\cite{mast3r} adds metric depth prediction with enhanced matching accuracy. To address dynamic content, MonST3R~\cite{zhang2025monstr} and Align3R~\cite{lu2024align3r} incorporate task-specific fine-tuning, while Easi3R~\cite{chen2025easi3r} enables dynamic reconstruction without auxiliary cues. Other models such as Stereo4D~\cite{jin2024stereo4d}, Uni4D~\cite{yao2025uni4d}, ZeroShot-MSF~\cite{Liang2025ZeroShotMSF}, Dynamic Point Maps~\cite{sucar2025dynamic}, D$^2$USt3R~\cite{han2025d}, MMP~\cite{park2025learning}, POMATO~\cite{zhang2025pomato}, BA-Track~\cite{chen2025back}, STORM~\cite{yang2024storm}, and DPV~\cite{sucar2025dynamicpointmapsversatile} jointly model motion and geometry. To move beyond stereo pairs, Spann3R~\cite{spann3r}, CUT3R~\cite{wang2025cut3r}, and Regist3R~\cite{liu2025regist3r} introduce memory and recurrence for longer sequences, while Fast3R~\cite{Yang_2025_Fast3R}, VGGT~\cite{wang2025vggt}, MASt3R-SfM~\cite{duisterhof2024mast3r}, MUSt3R~\cite{must3r}, Light3R-SfM~\cite{light3r_sfm}, and MV-DUSt3R+~\cite{tang2025mv} support scalable multi-view inference with outputs in global coordinates. Several of these models also support appearance modeling for view synthesis. Methods like LSM~\cite{fan2024largespatialmodelendtoend}, FLARE~\cite{zhang2025FLARE}, Splatt3R~\cite{smart2024splatt3r}, NoPoSplat~\cite{ye2024no}, SpatialSplat~\cite{sheng2025spatialsplat}, Matrix3D~\cite{lu2025matrix3d}, AnySplat~\cite{jiang2025anysplat}, VicaSplat~\cite{li2025vicasplat}, and PREF3R~\cite{chen2024pref3r} predict 3D Gaussians for high-quality rendering. Notably, LSM further enables language-guided semantic understanding, such as text-driven 3D semantic segmentation. 
Recent works also explore how to incorporate more available input conditions~\cite{jang2025pow3r}, and further extend GFMs to broader tasks such as surface reconstruction~\cite{raj2024spurfiessparsesurfacereconstruction}, plane reconstruction~\cite{liu2025towards}, SLAM~\cite{li2024_MegaSaM, vggtslam, mast3r_slam, wildgs_slam, slam3r, li2025hier}, end-to-end aerial-ground 3D mapping~\cite{vuong2025aerialmegadepth}, autonomous driving~\cite{fei2024driv3r,li2025rig3r}, human-scene reconstruction~\cite{muller2025reconstructing, liu2025joint}, and 3D spatial reasoning~\cite{fan2025vlm3r}.

\textbf{Stream 2: Diffusion-based 3D GFMs.}

Another line of work leverages diffusion models to reconstruct per-pixel 3D point maps, and in some cases, camera poses, through iterative denoising. 
DiffusionSfM~\cite{diffusionsfm} parameterizes scene geometry and cameras as pixel-wise ray origins and endpoints in a global frame and directly infers 3D scene geometry and camera poses from multi-view images. 
Extending this idea to video, recent methods build on video diffusion models to tackle the more complex task of dynamic 4D reconstruction, where temporal consistency and object motion introduce additional challenges.
Aether~\cite{team2025aether} generates depth and ray maps via a diffusion backbone. GeometryCrafter~\cite{xu2025geometrycrafter} introduces a VAE-based architecture with dual encoder-decoders to improve pointmap quality from videos. Geo4D~\cite{Geo4D} unifies multiple geometric modalities—including points, rays, and depths—for temporally coherent reconstruction from video. UniGeo~\cite{sun2025unigeo} finetunes video diffusion model to predict geometry attributes such as surface normals and coordinates.

\textbf{End-to-End Sparse SfM.}
There is also a parallel line of work on sparse 3D reconstruction in an end-to-end manner, including VGGSfM~\cite{wang2024vggsfm} and FlowMap~\cite{smith2024flowmap}, which aim to replace classical SfM pipelines. While promising, these methods primarily focus on sparse reconstruction and are not the focus of our benchmark, which centers on dense 3D geometry prediction.

\subsection{Monocular Image/Video Depth Estimators.}
Before the emergence of 3D geometric foundation models (GFMs), numerous models were developed to estimate monocular depth from single images or videos. MiDaS~\cite{ranftl2020towards} introduced affine-invariant supervision and training on mixed datasets for better zero-shot generalization. Depth Anything~\cite{yang2024depthanything} and its V2~\cite{depth_anything_v2} extended this framework using transformer architectures and large-scale unlabeled data for semi-supervised learning. 
Pioneered by Marigold~\cite{ke2024repurposing}, recent works~\cite{garcia2025fine,fu2024geowizard,gui2024depthfm,he2024lotus,pham2024sharpdepth} adapted pretrained diffusion models by finetuning only the U-Net on latent depth codes, converting depth maps to pseudo-RGB representations. To generalize to videos, temporal consistency was achieved through test-time optimization, memory modules, or stabilization networks~\cite{zhang2021consistent,yasarla2023mamo,wang2023neural,kopf2021robust,luo2020consistent}, and more recently by directly finetuning video diffusion models~\cite{hu2024depthcrafter,shao2024learning,yang2024depthanything}. While effective at producing dense depth, these models do not predict camera intrinsics or poses, and their affine-invariant outputs limit use in full 3D reconstruction.

Some methods explicitly address scale ambiguity and camera awareness. LeReS~\cite{yin2021learning, yin2022towards} incorporates 3D point cloud encoders to recover missing focal length and shift. UniDepth~\cite{piccinelli2024unidepth} decouples camera parameter prediction from depth estimation via a pseudo-spherical representation and self-prompting. DepthPro~\cite{bochkovskii2024depth} introduces a ViT-based model with a focal length prediction head, while MoGe~\cite{wang2024moge} uses affine-invariant point maps to reduce focal-distance ambiguity. Metric3D~\cite{yin2023metric3d} and its V2~\cite{hu2024metric3d} estimate metric depth using known camera parameters. Although these approaches improve monocular depth estimation, they can only process a single image and cannot be used to image collections or videos for 3D reconstruction directly.

Thus, we treat monocular depth estimators as valuable baselines for evaluating depth capabilities but do not consider them full GFMs in our benchmark.

\subsection{Prior Benchmarks in 3D Reconstruction.}
Classic 3D benchmarks such as DTU~\cite{dtu}, Tanks and Temples~\cite{tanks}, ETH3D~\cite{eth3d}, KITTI~\cite{kitti, kitti_ordometry}, and ScanNet~\cite{scannet, scannet++}, have significantly advanced traditional pipelines for multi-view stereo (MVS)~\cite{schoenberger2016mvs}, structure-from-motion (SfM)~\cite{schoenberger2016sfm}, and SLAM~\cite{kerl2013dense}. However, these benchmarks are typically task-specific and constrained to narrow domains: DTU focuses on object-centric tabletop scenes, KITTI on outdoor street driving, and ScanNet on indoor environments. Moreover, they are designed for systems that require intermediate inputs such as camera intrinsics, keypoint matches, or depth maps at inference time.

In contrast, 3D geometric foundation models (GFMs) operate in an end-to-end manner, predicting geometry directly from RGB inputs without relying on such precomputed representations. While existing benchmarks still offer valuable ground-truth data for evaluation, they are not sufficient to assess GFM performance across diverse capture conditions and tasks.

To address these limitations, our benchmark complements prior efforts by providing a unified evaluation framework that examines both the effectiveness and efficiency of modern GFMs. For effectiveness, it spans five core tasks: sparse-view and video-based depth estimation, extremely sparse and dense 3D reconstruction, relative pose estimation, and novel view synthesis. It also expands domain coverage to include dynamic real-world scenes, aerial drone footage, and challenging air–ground paired trajectories. For efficiency, we benchmark inference latency and peak memory usage under varying input scales. Together, these components form a comprehensive testbed for evaluating the generalization capabilities and deployment readiness of today’s 3D geometric foundation models.

\section{Benchmark Details}

Tab.\ref{tab:dataset_process} summarizes the datasets and preprocessing procedures used in our benchmark. 
Detailed evaluation protocols for sparse-view and video depth estimation are provided Sec.~\ref{apx:sparse_depth} and Sec.~\ref{apx:video_depth} respectively. Implementation details, including model input and resolution setup, are described in Sec.~\ref{apx:imple_detail}.

\begin{table}[htbp] 
\centering
\caption{\textbf{Summary of Benchmark Tasks, Datasets, Number of Scenes, and Preprocessing}.}
\label{tab:dataset_process}
\resizebox{\textwidth}{!}{%
\begin{tabular}{cccp{7cm}} 
\toprule
\textbf{Task} & \textbf{Dataset} & \textbf{\# of Scenes} & \textbf{Dataset Preprocessing} \\
\midrule

\multirow{5}{*}{Sparse-View Depth Estimation} & DTU & 110 & \multirow{5}{*}{Following RobustMVD protocol} \\ 
& ScanNet & 200 & \\
& KITTI & 93 & \\
& ETH3D & 104 & \\
& Tanks and Temples & 69 & \\
\midrule

\multirow{6}{*}{Video Depth Estimation} & Bonn & 5 & Selected frames 30-140, sampled with stride=2 \\
& TUM Dynamics & 8 & Sampled with stride=30 \\
& KITTI & 13 & First 110 frames, sampled with stride=2 \\
& PointOdyssey val & 15 & First 110 frames, sampled with stride=2  \\
& Syndrone & 8 & First 100 frames, sampled with stride=2  \\
& Sintel & 14 & All frames used \\
\midrule

\multirow{11}{*}{Multi-View Relative Pose Estimation} & CO3Dv2 & 2511 & Randomly selected 10 frames \\
& RealEstate10k & 1611 & Randomly selected 10 frames \\
& ScanNet-eval & 89 & First 1200 frames, sampled with stride=30 \\
& Bonn & 5 & Selected frames 30-140, sampled with stride=2 \\
& TUM Dynamics & 8 & Sampled with stride=30 \\
& KITTI Ordometry & 11 & First 110 frames, sampled with stride=2 \\
& Sintel & 14 & All frames used \\
& ADT & 8 & Sampled with stride=10 \\
& ACID & 1500 & Randomly selected 10 frames \\
& Syndrone & 8 & First 100 frames, sampled with stride=2\\
& ULTRRA & 4 & Pairs constructed following AerialMegaDepth \\
\midrule

\multirow{5}{*}{\makecell{Multi-View 3D Reconstruction \\ (Extremely Sparse)}} & DTU & 22 & Sampled with stride=16 \\
& 7-Scenes & 18 & Sampled with stride=200 \\
& NRGBD & 9 & Sampled with stride=500 \\
& ScanNet & 20 & First 1200 frames, sampled with stride=300 \\
& TUM RGBD & 11 & First 1200 frames, sampled with stride=300 \\
\midrule

\multirow{5}{*}{\makecell{Multi-View 3D Reconstruction \\ (Dense)}} & DTU & 22 & Sampled with stride=5 \\
& 7-Scenes & 18 & Sampled with stride=30 \\
& NRGBD & 9 & Sampled with stride=40 \\
& ScanNet & 20 & First 1200 frames, sampled with stride=30 \\
& TUM RGBD & 11 & First 1200 frames, sampled with stride=30 \\
\midrule

\multirow{4}{*}{Novel View Synthesis} & DTU & 16 & \multirow{4}{*}{Following NoPoSplat protocol} \\
& RealEstate10k & 34 & \\
& ScanNet++ & 50 & \\
& ACID & 11 & \\
\bottomrule
\end{tabular}%
}
\end{table}

\subsection{Sparse-View Depth Estimation}\label{apx:sparse_depth}

\textbf{Additional Evaluation Details:}
Depth maps are extracted from the z-coordinate of predicted pointmaps after projection into the camera coordinate system. For models that output both camera- and world-coordinate pointmaps, we use the pointmaps in the world coordinate system to ensure consistency.

For GFMs requiring global alignment to handle multi-view inputs, we extract depth maps after alignment. Specifically, DUSt3R and MASt3R perform global alignment using complete scene graphs, while MonST3R adopts a sliding-window strategy. However, MonST3R’s flow loss occasionally produces invalid values (e.g., NaN), particularly on static datasets. In such cases, we disable the flow loss module to maintain evaluation integrity.

To isolate geometry quality from camera estimation errors, we use ground-truth camera intrinsics and poses for all models. To fairly compare models with different output scales, we report results under both normalized and metric settings: 1) For normalized-scale models, we apply per-view median alignment, using the ratio between predicted and ground-truth median depths. 2) For metric-scale models, we report two variants: (i) raw predictions, and (ii) the same median-aligned depths used for normalized models, enabling consistent comparisons.

To control for variations due to input view selection, we follow the quasi-optimal source view strategy proposed in RobustMVD~\cite{schroppel2022benchmark}. If global alignment fails to converge with quasi-optimal views, we fall back to using the nearest neighboring views to ensure evaluation continuity.

\subsection{Video Depth Estimation}
\label{apx:video_depth}

\textbf{More Evaluation Details:} We evaluate both normalized- and metric-scale models. For normalized-scale models, we apply a scale-and-shift alignment to ground truth before computing metrics. For metric-scale models, we report two variants: (i) using raw predictions, and (ii) applying the same scale-and-shift alignment for fair comparison. To consistently evaluate temporal coherence, we follow MonST3R and compute a single global scale and shift per video sequence—rather than per frame—before metric evaluation. This sequence-level alignment ensures that temporal consistency is not biased by framewise normalization.



\subsection{Implementation Details}\label{apx:imple_detail}
GFMs vary in their input resolution requirements—for example, VGGT requires both height and width to be divisible by 14; DUSt3R and MASt3R typically use $512{\times}384$; Spann3R uses $224{\times}224$. To preserve each model’s intended behavior, we adopt its native input resolution. However, for fair comparison across tasks, we standardize output resolution: we use the original resolution for sparse depth estimation and $512{\times}384$ for all other tasks. For diffusion-based methods, which require a fixed number of input frames (e.g., Geo4D uses 16 frames), we pad shorter sequences by repeating the last frame to meet the input requirement. We omit 3D reconstruction evaluation for all diffusion models, and pose estimation for GeometryCrafter, as these tasks are not natively supported by those architectures.

\section{More Quantitative Results}

\subsection{Multi-View Relative Pose Estimation}
\label{apx:multi_view_pose_more}

\textbf{Per-Dataset Breakdown:}
We evaluate pose estimation across 11 datasets and group them into six representative settings: \textit{In-Distribution} (CO3Dv2), \textit{Long Sequence} (ScanNet, ADT, and TUM-Dynamics), \textit{Street Driving} (KITTI Odometry), \textit{Indoor-Outdoor} (Bonn, Sintel, and RealEstate10K), \textit{Drone} (ACID, Syndrone), and \textit{Air-Ground} (ULTRRA Challenge). In Tab.4 of the main paper, we report scene-level averages for each setting to mitigate the effect of varying metric scales across datasets. For finer-grained comparison, we include a per-dataset breakdown for the \textit{Long Sequence}, \textit{Indoor-Outdoor}, and \textit{Drone} categories in Tab.~\ref{tab:perdataset_pose}.

\begin{table*}[ht]
\centering
\scriptsize
\caption{\textbf{Per-Dataset Breakdown of Multi-view Relative Pose Estimation across Diverse Scenarios.} We report ATE (↓), RPE translation (↓), and RPE rotation (↓) for each dataset.}
\label{tab:perdataset_pose}
\vspace{-3pt}
\begin{subtable}{\linewidth}
\centering
\caption{\textbf{Long Sequence Scenarios (Blue)}}
\label{tab:long_sequence}
\resizebox{\linewidth}{!}{
\begin{tabular}{lccccccccc}
\toprule
\multirow{2}{*}{\textbf{Method}}
& \multicolumn{3}{c}{\cellcolor{blue!15}\textbf{ScanNet}}
& \multicolumn{3}{c}{\cellcolor{blue!15}\textbf{ADT}}
& \multicolumn{3}{c}{\cellcolor{blue!15}\textbf{TUM-Dynamics}} \\
& \cellcolor{blue!15}ATE ↓ & \cellcolor{blue!15}RPE$_\text{trans}$ ↓ & \cellcolor{blue!15}RPE$_\text{rot}$ ↓& \cellcolor{blue!15}ATE ↓ & \cellcolor{blue!15}RPE$_\text{trans}$ ↓ & \cellcolor{blue!15}RPE$_\text{rot}$ ↓
& \cellcolor{blue!15}ATE ↓ & \cellcolor{blue!15}RPE$_\text{trans}$ ↓ & \cellcolor{blue!15}RPE$_\text{rot}$ ↓ \\
\midrule
DUSt3R/LSM & \cellcolor{top3}0.098 & \cellcolor{top3}0.071 & \cellcolor{top2}2.072 & \cellcolor{top3}0.653 & 0.454 & \cellcolor{top3}5.651 & 0.077 & 0.089 & \cellcolor{top3}2.709\\
MASt3R& \cellcolor{top2}0.086 & \cellcolor{top2}0.067 & \cellcolor{top3}2.112 & 0.663 & 0.439 & \cellcolor{top1}5.628 & 0.098 & 0.102 & 8.790\\
Spann3R& 0.295 & 0.144 & 3.764 & \cellcolor{top1}0.520 & 0.495 & 6.147 & \cellcolor{top2}0.064 & \cellcolor{top2}0.054 & \cellcolor{top2}1.564\\
CUT3R& 0.151 & 0.112 & 3.917 & 0.675 & \cellcolor{top3}0.436 & \cellcolor{top2}5.635 & \cellcolor{top3}0.065 & \cellcolor{top3}0.069 & 9.474\\
VGGT & \cellcolor{top1}0.070 & \cellcolor{top1}0.060 & \cellcolor{top1}1.249 & 0.694 & 0.439 & 5.673 & \cellcolor{top1}0.015 & \cellcolor{top1}0.018 & \cellcolor{top1}0.586\\
Fast3R & 0.513 & 0.405 & 26.286 & 0.737 & 0.508 & 9.720 & 0.106 & 0.116 & 9.424\\
MonST3R & 0.463 & 0.258 & 12.947 & \cellcolor{top2}0.535 & 0.739 & 10.228 & 0.192 & 0.146 & 13.957\\
Align3R & 0.431 & 0.206 & 9.487 & 0.674 & 0.440 & 6.256 & 0.103 & 0.086 & 11.965\\
Easi3R & 0.135 & 0.074 & 2.569 & 0.687 & 0.447 & 5.830 & 0.100 & 0.086 & 3.282\\
Geo4D & 0.428 & 0.156 & 10.486 & 0.793 & \cellcolor{top2}0.423 & 9.408 & 0.176 & 0.139 & 12.593\\
Aether  & 0.681 & 0.283 & 15.555 & 0.704 & \cellcolor{top1}0.328 & 8.717 & 0.174 & 0.111 & 12.535 \\
\bottomrule
\end{tabular}
}
\end{subtable}

\begin{subtable}{\linewidth}
\centering
\caption{\textbf{Indoor-Outdoor Scene (Orange)}}
\label{tab:indoor_outdoor}
\resizebox{\linewidth}{!}{
\begin{tabular}{lccccccccc}
\toprule
\multirow{2}{*}{\textbf{Method}}
& \multicolumn{3}{c}{\cellcolor{orange!15}\textbf{Bonn}}
& \multicolumn{3}{c}{\cellcolor{orange!15}\textbf{Sintel}}
& \multicolumn{3}{c}{\cellcolor{orange!15}\textbf{RealEstate10k}} \\
& \cellcolor{orange!15}ATE ↓ & \cellcolor{orange!15}RPE$_\text{trans}$ ↓ & \cellcolor{orange!15}RPE$_\text{rot}$ ↓
& \cellcolor{orange!15}ATE ↓ & \cellcolor{orange!15}RPE$_\text{trans}$ ↓ & \cellcolor{orange!15}RPE$_\text{rot}$ ↓
& \cellcolor{orange!15}ATE ↓ & \cellcolor{orange!15}RPE$_\text{trans}$ ↓ & \cellcolor{orange!15}RPE$_\text{rot}$ ↓ \\
\midrule
DUSt3R/LSM  & 0.026 & 0.022 & 1.259 & 0.355 & 0.202 & 13.740 & 0.075 & 0.562 & 1.553\\
MASt3R & 0.022 & 0.022 & 1.245 & 0.340 & 0.291 & 5.977 & \cellcolor{top2}0.056 & 0.563 & 1.265\\
Spann3R & 0.041 & 0.015 & 1.721 & 0.329 & 0.114 & 4.114 & 0.081 & 0.102 & 1.272\\
CUT3R & 0.033 & 0.015 & 1.150 & 0.209 & 0.071 & 0.634 & \cellcolor{top1}0.031 & \cellcolor{top1}0.039 & \cellcolor{top1}0.497\\
VGGT & \cellcolor{top1}0.013 & 0.016 & 1.118 & \cellcolor{top3}0.172 & \cellcolor{top3}0.062 & \cellcolor{top1}0.471 & \cellcolor{top3}0.061 & 0.111 & \cellcolor{top2}0.579\\
Fast3R   &0.034 & 0.035 & 1.594 & 0.267 & 0.209 & 14.342 & 0.110 & 0.170 & 1.911\\
MonST3R & 0.023 & \cellcolor{top3}0.013 & \cellcolor{top2}1.105 & \cellcolor{top1}0.107 & \cellcolor{top1}0.039 & 0.664 & 0.098 & 0.154 & \cellcolor{top3}0.831\\
Align3R & 0.022 & \cellcolor{top3}0.013 & 1.115 & 0.542 & 0.150 & \cellcolor{top3}0.602 & 0.072 & \cellcolor{top2}0.091 & 1.088\\
Easi3R & \cellcolor{top2}0.019 & 0.016 & 1.308 & 0.370 & 0.212 & 13.726 & 0.073 & \cellcolor{top3}0.093 & 1.254\\
Geo4D & 0.026 & \cellcolor{top1}0.012 & \cellcolor{top3}1.113 & 0.179 & 0.064 & \cellcolor{top2}0.515 & 0.578 & 0.477 & 3.816\\
Aether & \cellcolor{top2}0.019 & \cellcolor{top1}0.012 & \cellcolor{top1}0.863 & \cellcolor{top2}0.158 & \cellcolor{top2}0.046 & 0.632 & 0.195 & 0.123 & 1.621 \\
\bottomrule
\end{tabular}
}
\end{subtable}
\vspace{4pt}

\begin{subtable}{\linewidth}
\centering
\caption{\textbf{Drone Scene (Cyan)}}
\label{tab:drone}
\resizebox{0.7\linewidth}{!}{
\begin{tabular}{lcccccc}
\toprule
\multirow{2}{*}{\textbf{Method}}
& \multicolumn{3}{c}{\cellcolor{cyan!15}\textbf{ACID}}
& \multicolumn{3}{c}{\cellcolor{cyan!15}\textbf{Syndrone}} \\
& \cellcolor{cyan!15}ATE ↓ & \cellcolor{cyan!15}RPE$_\text{trans}$ ↓ & \cellcolor{cyan!15}RPE$_\text{rot}$ ↓& \cellcolor{cyan!15}ATE ↓ & \cellcolor{cyan!15}RPE$_\text{trans}$ ↓ & \cellcolor{cyan!15}RPE$_\text{rot}$ ↓ \\
\midrule
DUSt3R/LSM & 0.124 & 0.368 & 2.849 & \cellcolor{top3}0.535 & 2.565 & 0.300 \\
MASt3R & 0.129 & 0.364 & 2.614 & \cellcolor{top2}0.324 & 2.542 & \cellcolor{top3}0.214\\
Spann3R& \cellcolor{top2}0.108 & 0.136 & 1.484 & 1.817 & 2.706 & 1.398 \\
CUT3R& \cellcolor{top1}0.062 & \cellcolor{top1}0.077 & \cellcolor{top3}0.917 & 1.684 & 2.565 & 0.311\\
VGGT & 0.280 & 0.450 & \cellcolor{top2}0.806 & \cellcolor{top1}0.206 & \cellcolor{top3}2.532 & \cellcolor{top1}0.109\\
Fast3R & 0.431 & 0.505 & 1.984 & 1.537 & 3.119 & 1.026\\
MonST3R& 0.321 & 0.493 & 1.519 & 2.966 & 2.563 & 0.594\\
Align3R& 0.144 & 0.167 & 0.981 & 1.208 & \cellcolor{top2}2.528 & \cellcolor{top2}0.133\\
Easi3R & \cellcolor{top3}0.109 & \cellcolor{top3}0.125 & 1.740 & 2.012 & 2.560 & 0.256\\
Geo4D& 0.376 & 0.317 & 1.400 & 1.795 & 2.550 & 0.434\\
Aether& 0.146 & \cellcolor{top2}0.088 & \cellcolor{top1}0.796 & 1.415 & \cellcolor{top1}1.759 & 0.727\\
\bottomrule
\end{tabular}
}
\end{subtable}

{\tiny
\begin{tabular}{@{}l@{}}
\colorbox{blue!15}{\strut\textbf{Blue}}: Long Sequence \quad
\colorbox{orange!15}{\strut\textbf{Orange}}: Indoor-Outdoor Scene \quad
\colorbox{cyan!15}{\strut\textbf{Cyan}}: Drone
\end{tabular}
}
\vspace{-6pt}
\end{table*}

We observe several trends in the results: 1) VGGT demonstrates the best generalization, dominating the long sequence scenarios and  drone footage. 2) In addition to DUSt3R and MASt3R, methods designed for online registration, such as Spann3R and CUT3R, also perform well on long video sequences.
3) Among diffusion-based models, Aether consistently ranks among the top performers, while Geo4D shows strong results primarily in indoor-outdoor scenes. 

\textbf{Additional Comparison on the ULTRRA Challenge:}
Our pose estimation metrics are computed using Sim(3) Umeyama alignment between predicted and ground-truth trajectories. However, in the ULTRRA dataset, the aerial and ground trajectories are reconstructed in separate coordinate frames. Despite being calibrated with RTK-corrected GPS, the absence of a shared reference system makes a single global alignment infeasible, rendering ATE inapplicable. As a result, we omit ATE and only report RPE-trans and RPE-rot in Tab.4 of the main paper. Notably, both metrics are significantly worse than in all other settings, making it difficult to discern which GFMs perform well in this challenging domain. 

To supplement these results, in Tab.~\ref{tab:ultrra}, we report RTA@$\tau$ and RRA@$\tau$: the percentage of camera pairs with relative translation or rotation error below threshold $\tau$. These metrics better reflect performance under the unique challenges of this setting. Diffusion-based methods (e.g., Geo4D, Aether) are excluded from this evaluation, as they require fixed-length input sequences (e.g., 16 and 41 frames respectively), making them incompatible with the pairwise-view setup in ULTRRA.

\begin{table}[ht]
\centering
\caption{\textbf{Camera Rotation and Translation Accuracy on ULTRRA challenge} (excluding fine-tuned models from top-3 ranking).}
\label{tab:ultrra}
\resizebox{0.8\linewidth}{!}{
\begin{tabular}{lcccccc}
\toprule
\multirow{2}{*}{\textbf{Method}} & \multicolumn{3}{c}{\textbf{Camera Rotation Accuracy}} & \multicolumn{3}{c}{\textbf{Camera Translation Accuracy}} \\
\cmidrule(lr){2-4}\cmidrule(lr){5-7}
& RRA@5 ↑& RRA@10 ↑& RRA@15 ↑& RTA@5 ↑& RTA@10 ↑& RTA@15 ↑\\
\midrule
DUSt3R \tiny{(ft AerialMegaDepth)} & 55.96 & 71.25 & 76.15 & 46.48 &68.20 & 72.78 \\
MASt3R \tiny{(ft AerialMegaDepth)}& 49.54 & 66.36 & 72.48 & 42.51 & 63.30 & 69.11 \\ \hline
DUSt3R& \cellcolor{top3}5.50 & 8.56 & 9.79 & \cellcolor{top2}2.45 & \cellcolor{top3}6.12 & 8.26 \\
MASt3R& 3.06 & 3.98 & 5.50 & \cellcolor{top3}1.83 & 3.67 & 4.89 \\
VGGT& \cellcolor{top2}6.73 & 10.09 & 13.15 & 1.53 & 4.89 & 6.42 \\
Fast3R& \cellcolor{top3}5.50 & \cellcolor{top3}12.84 & \cellcolor{top2}22.02 & \cellcolor{top2}2.45 & \cellcolor{top2}7.34 & \cellcolor{top2}11.01 \\
Spann3R & \cellcolor{top1}9.79 & \cellcolor{top1}26.30 & \cellcolor{top1}38.53 & \cellcolor{top1}4.28 & \cellcolor{top1}8.87 & \cellcolor{top1}15.90 \\
MonST3R & 1.83 & 2.75 & 3.67 & 0.31 & 1.83 & 2.75 \\
CUT3R & 3.67 & \cellcolor{top2}16.51 & \cellcolor{top2}22.02 & 0.31 & 2.45 & 5.20 \\
Align3R & 4.11 & 9.89 & 12.08 & 1.71 & 3.85 & 6.27 \\
Easi3R& 4.70 & 10.26 & \cellcolor{top3}13.68 & 1.28 & 5.13 & \cellcolor{top3}8.97 \\
\bottomrule
\end{tabular}
}
\end{table}

From these results, we observe: 1) Without fine-tuning on datasets like AerialMegaDepth, Spann3R achieves the best accuracy, followed closely by Fast3R. 2) Fine-tuning clearly provides substantial performance gains for this challenging out-of-distribution scenario, highlighting the importance of domain-specific adaptation to improve the domain generalization of GFMs, as mentioned in our findings.

\subsection{Novel View Synthesis}
\label{apx:nvs_more}

For DUSt3R variants equipped with appearance modeling (i.e., LSM, Splatt3R, NoPoSplat, and FLARE), training is typically performed on narrow domain-specific datasets, for example, Splatt3R is trained only on ScanNet++, and NoPoSplat on ACID or RealEstate10K. Unlike methods such as PixelSplat~\cite{charatan2023pixelsplat} and MVSplat~\cite{chen2024mvsplat}, these models are all pose-free: given a pair of images, they directly predict per-pixel 3D Gaussians without requiring ground-truth camera intrinsics or extrinsics. The resulting scenes are scale-invariant, meaning ground-truth novel-view poses cannot be directly applied at inference time. To address this, NoPoSplat and FLARE optimize the test-time novel pose via photometric losses (PSNR + LPIPS), while LSM and Splatt3R rescale the ground-truth pose using the relative scale between predicted and ground-truth pointmaps.

Since not all of our selected datasets provide ground-truth depth, we apply test-time pose optimization for all methods to ensure a fair and consistent comparison. Splatt3R is not included in this comparison because its evaluation code was not publicly released at the time of writing. All evaluations are conducted at a resolution of $256{\times}256$, and the results are summarized in Tab.~\ref{tab:nvs}. For NoPoSplat, which provides two separately trained models on ACID and RealEstate10K, we follow the official instructions and use the corresponding model for each respective test set and use the model trained on RealEstate10K for all other datasets by default to maintain consistency across evaluations.

\begin{table}[ht]
\centering
\scriptsize
\vspace{-5pt}
\caption{\small \textbf{Comparison on Novel View Synthesis}. We report PSNR (↑), SSIM (↑), and LPIPS (↓).}
\label{tab:nvs}
\resizebox{0.98\linewidth}{!}{
\begin{tabular}{lcccccccccccc}
\toprule
\multirow{2}{*}{\textbf{Method}}
& \multicolumn{3}{c}{\cellcolor{gray!20}\textbf{DTU}}
& \multicolumn{3}{c}{\cellcolor{blue!15}\textbf{RealEstate10k}}
& \multicolumn{3}{c}{\cellcolor{blue!15}\textbf{ScanNet++}}
& \multicolumn{3}{c}{\cellcolor{cyan!15}\textbf{ACID}} \\
& \cellcolor{gray!20}PSNR ↑& \cellcolor{gray!20}SSIM ↑& \cellcolor{gray!20}LPIPS ↓
& \cellcolor{blue!15}PSNR ↑& \cellcolor{blue!15}SSIM ↑& \cellcolor{blue!15}LPIPS ↓
& \cellcolor{blue!15}PSNR ↑& \cellcolor{blue!15}SSIM ↑& \cellcolor{blue!15}LPIPS ↓
& \cellcolor{cyan!15}PSNR ↑& \cellcolor{cyan!15}SSIM ↑& \cellcolor{cyan!15}LPIPS ↓\\
\midrule
LSM& \cellcolor{top2}17.38 & \cellcolor{top2}0.6274 & \cellcolor{top3}0.3198 & \cellcolor{top3}18.92 & \cellcolor{top3}0.6677 & \cellcolor{top3}0.3643 & \cellcolor{top3}17.12 & \cellcolor{top3}0.6860 & \cellcolor{top3}0.3887 & \cellcolor{top3}20.46 & \cellcolor{top3}0.6160 & \cellcolor{top3}0.3822\\
NoPoSplat& \cellcolor{top1}17.91 & \cellcolor{top1}0.6306 & \cellcolor{top1}0.2810 & \cellcolor{top1}24.53 & \cellcolor{top1}0.8450 & \cellcolor{top1}0.1634 & \cellcolor{top2}22.15 & \cellcolor{top2}0.7988 & \cellcolor{top2}0.2359 & \cellcolor{top1}25.35 & \cellcolor{top1}0.7774 & \cellcolor{top1}0.1875\\
FLARE & \cellcolor{top3}17.01 & \cellcolor{top3}0.5672 & \cellcolor{top2}0.2901 & \cellcolor{top2}22.15 & \cellcolor{top2}0.7126 & \cellcolor{top2}0.2363 & \cellcolor{top1}23.19 & \cellcolor{top1}0.8117 & \cellcolor{top1}0.2201 & \cellcolor{top2}22.44 & \cellcolor{top2}0.6229 & \cellcolor{top2}0.2818\\
\bottomrule
\end{tabular}
}
{\tiny
\begin{tabular}{@{}l@{}}
\colorbox{gray!20}{\strut\textbf{Gray}}: Object-Centric \quad
\colorbox{blue!15}{\strut\textbf{Blue}}: Indoor Scene \quad
\colorbox{cyan!15}{\strut\textbf{Cyan}}: Drone Scene
\end{tabular}
}
\vspace{-10pt}
\end{table}

From the table, we could observe that: 1) There is a clear performance gap between in-domain and out-of-domain settings. For instance, NoPoSplat performs best on ACID and RealEstate10k (datasets it was trained on) but its performance drops significantly on DTU and ScanNet++. Similarly, FLARE, trained on ScanNet++, performs best on its native test set. 2) In general, all methods perform worse on the object-centric DTU dataset, likely because their training data consists primarily of scene-level datasets, limiting their ability to generalize to object-level geometry and appearance. 3) LSM underperforms across all settings, which we attribute to instability in novel view pose optimization. Specifically, LSM relies on scaling ground-truth poses using the ratio between predicted and ground-truth pointmap scales. When the predicted pointmap scale is inaccurate or misaligned, it can lead to suboptimal pose estimation and degraded test-time performance.

More qualitative results and discussion are provided in Sec.~\ref{apx:nvs_visual}. These findings highlight both the promise and current limitations of appearance-aware GFMs when evaluated under cross-domain generalization.






\section{Qualitative Results}

\subsection{Extremely Sparse-View 3D Reconstruction}\label{apx:sparse_3d_recon_visual}

\begin{figure*}[!htb]
    \centering
    \includegraphics[width=0.9\linewidth]{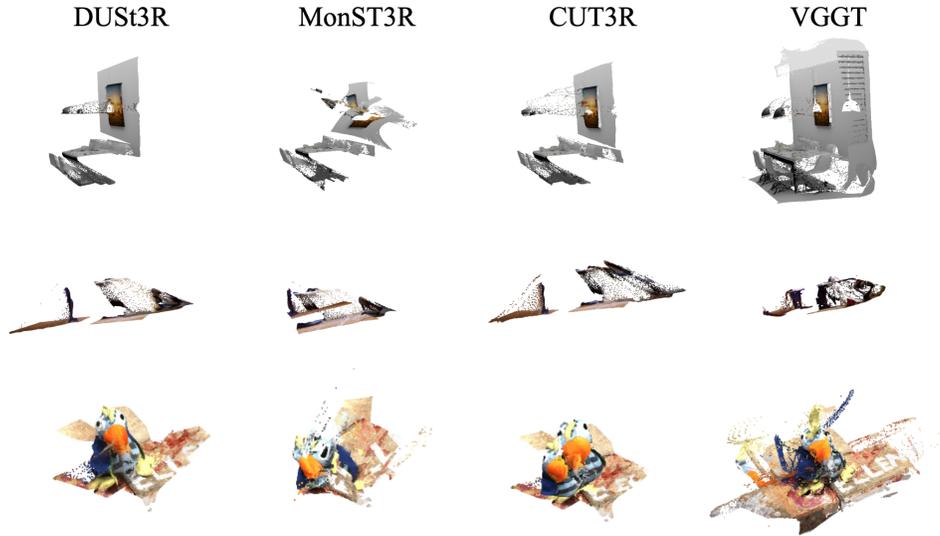}
    \vspace{-3mm}
    \caption{\textbf{Qualitative Comparison of Extremely Sparse-View 3D Reconstruction}. From top to bottom are scenes from NRGBD, TUM RGBD, and DTU. }
    \label{fig:3d_recon}
    \vspace{-10pt}
\end{figure*}

As noted in Tab.5 of the main paper, VGGT, CUT3R, FLARE, DUSt3R, and MASt3R perform well in extremely sparse-view 3D reconstruction, while models such as MonST3R, Align3R, and Easi3R show noticeably lower performance (with MonST3R included as a representative example). To illustrate these differences, we visualize reconstructions from VGGT, CUT3R, DUSt3R, and MonST3R in Fig.~\ref{fig:3d_recon}. We observe the following: (1) For image sets with minimal or no overlap, VGGT consistently reconstructs clean and coherent 3D structures, whereas MonST3R often produces outliers. (2) DUSt3R performs especially well on the DTU dataset, which aligns with its strong quantitative results reported in Tab.5 in main draft.

\subsection{Novel View Synthesis}\label{apx:nvs_visual}

\begin{figure*}[!htb]
    \centering
    \includegraphics[width= 0.9\linewidth]{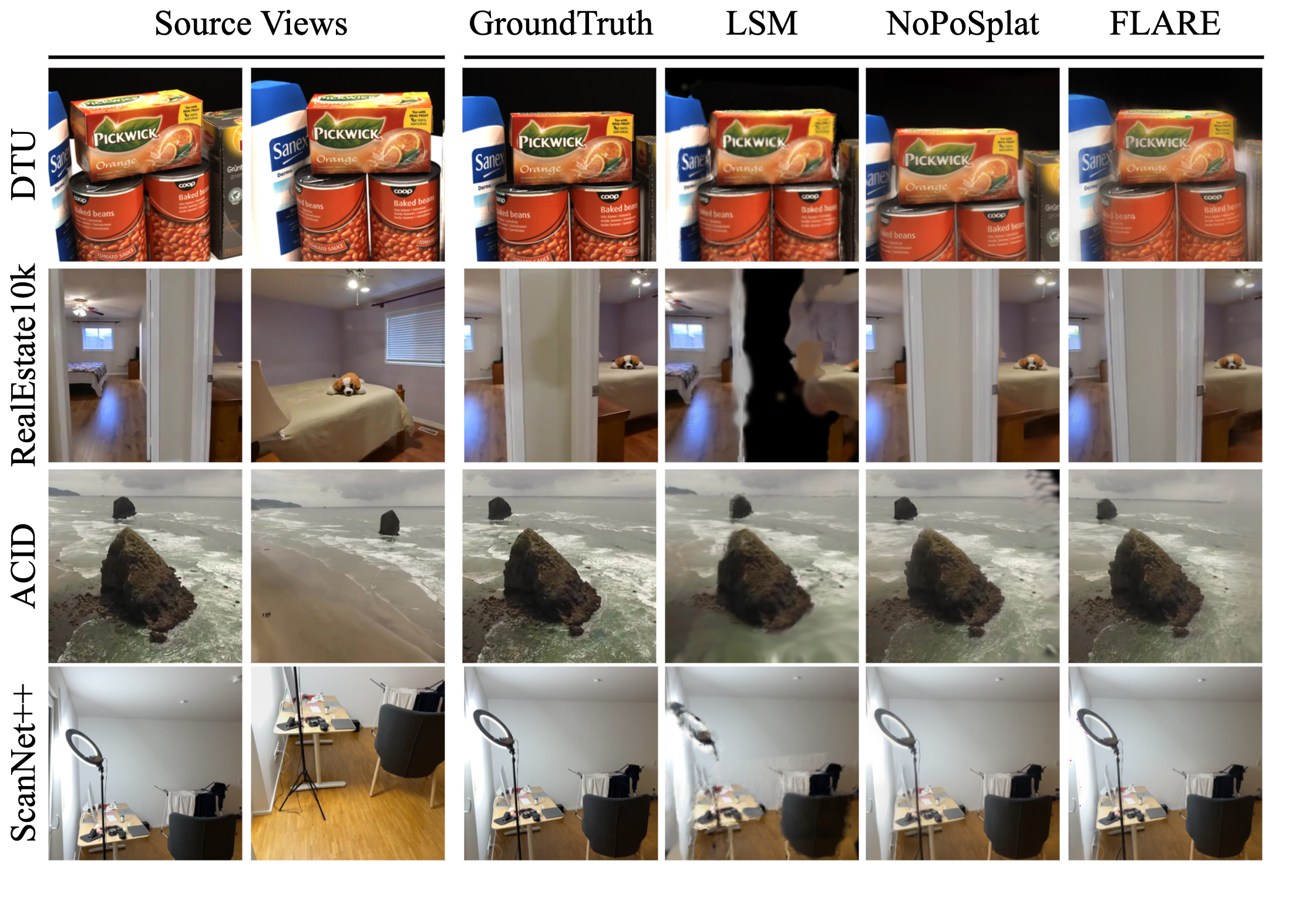}
    \vspace{-3mm}
    \caption{\textbf{Qualitative Comparison of Novel View Synthesis} given two source views. Rendering resolution is 256$\times$256.}
    \label{fig:nvs}
    \vspace{-16pt}
\end{figure*}

Novel view synthesis results on DTU, ScanNet++, RealEstate10K, and ACID are shown in Fig.~\ref{fig:nvs}. We observe the following: 1) LSM often produces visible holes in the synthesized views (e.g., the wall in RealEstate10K), and struggles with fine-grained structures, (e.g., thin floor lamps in ScanNet++), possibly due to the erroneous pointmap prediction. 2) When using test-time pose optimization, the optimized pose may still fail to perfectly align with ground-truth, especially on datasets like DTU. This misalignment is evident across all methods and helps explain their relatively lower performance on DTU in Tab.~\ref{tab:nvs}.

\subsection{Video Depth Estimation}\label{apx:video_depth_visual}

\begin{figure*}[!htb]
    \centering
    \includegraphics[width=\linewidth]{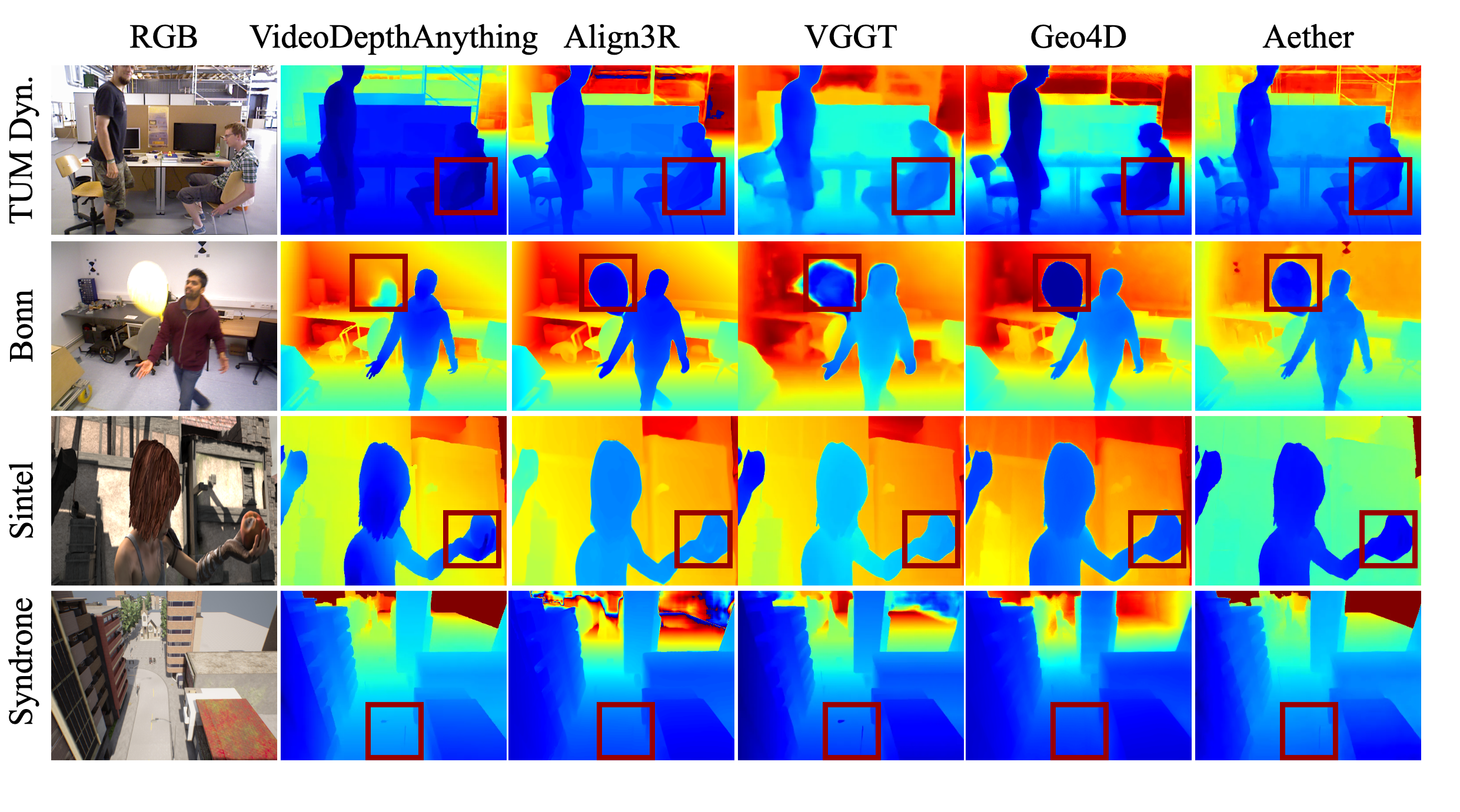}
    \vspace{-3mm}
    \caption{\textbf{Qualitative Comparison of Video Depth Estimation}.}
    \label{fig:videodepth}
    \vspace{-10pt}
\end{figure*}

In Tab.3 in the main draft, we observe that current GFMs perform competitively, and in some cases surpass, specialized video depth models such as DepthCrafter and Marigold. In Fig.~\ref{fig:videodepth}, we visualize results from four top-performing GFMs (VGGT, Align3R, Geo4D, and Aether), alongside the baseline VideoDepthAnything.

We observe that: 1) VGGT captures fine-grained geometric relationships, such as the relative depth between the chair and arm in TUM Dynamics, relative depth between hand and apple in Sintel, or the presence of a streetlamp in Syndrone. Although its predictions may appear slightly blurred at object boundaries, its depth structure is semantically accurate. 2) Align3R produces sharp object silhouettes (e.g., monitor edges, chair legs) but exhibits less smoothness in distant regions, particularly in outdoor settings like Syndrone. 3) Geo4D and Aether (diffusion-based GFMs) yield crisp object contours but occasionally miss finer details, likely due to their iterative denoising formulation.

These qualitative results highlight that GFMs, despite not being explicitly designed for video depth, can rival or outperform task-specific approaches across a range of scenes.

\section{Future Directions}

Geometric foundation models (GFMs) represent a promising foundation for a wide range of emerging research directions beyond traditional 3D reconstruction tasks. One important future trajectory is their use as a \textbf{data generator}. For example, recent works like \cite{khazatsky2024droid, li2025sekai} leverage GFMs' ability to calibrate cameras or predict consistent 3D geometry from arbitrary data sources (e.g., Internet videos, or egocentric streams) to produce high-quality 3D supervision for downstream tasks. Such synthetic 3D supervision could reduce reliance on expensive ground-truth annotations, especially for tasks where dense 3D labels are scarce or difficult to obtain.

Another compelling direction lies in using GFMs as \textbf{spatial priors or encoders} within multimodal foundation models. For instance, recent work like~\cite{fan2025vlm3r, wu2025spatial, huang2025mllms, caothousands} demonstrates how geometric representations can enhance visual-language models (VLMs) by enabling spatial reasoning, 3D-aware grounding, and multimodal alignment across views. Incorporating geometric understanding into large multimodal systems could improve their ability to reason about spatial relationships, perform grounded generation, and interpret complex physical scenes from limited visual input.

Looking ahead, we envision GFMs playing a central role in future embodied intelligence systems, serving as a unified backbone for spatial understanding, environment simulation, and interaction planning in real-world 3D environments.
\end{document}